\pdfoutput=1

\documentclass[11pt]{article}
\usepackage[]{acl}
\usepackage{times} 
\usepackage{helvet} 
\usepackage{courier} 
\usepackage{graphicx} 
\usepackage{graphicx} 
\usepackage{natbib} 
\usepackage{caption} 
\usepackage{dirtytalk}
\usepackage{times}
\usepackage{latexsym}
\usepackage{graphicx}

\usepackage[T1]{fontenc}

\usepackage{latexsym}
\usepackage{graphicx}
\usepackage{amsmath, makecell}
\usepackage{amssymb}
\usepackage{booktabs}
\usepackage{mathtools}
\usepackage{multirow}
\usepackage{float}
\usepackage{placeins}
\usepackage[accsupp]{axessibility}
\usepackage{nopageno}
\usepackage{todonotes}
\usepackage{microtype}

\title{An Audit on the Perspectives and Challenges of Hallucinations in NLP}

\author{%
Pranav Narayanan Venkit\textsuperscript{$1$} \enspace Tatiana Chakravorti\textsuperscript{$1$} \enspace Vipul Gupta\textsuperscript{$4$} \enspace Heidi Biggs\textsuperscript{$3$} \\ \textbf{ \enspace Mukund Srinath\textsuperscript{$1$} \enspace 
Koustava Goswami\textsuperscript{$2$}\enspace Sarah Rajtmajer\textsuperscript{$1$} \enspace Shomir Wilson\textsuperscript{$1$}} \\
{\textsuperscript{$1$} {College of Information Sciences and Technology, Pennsylvania State University}}  \quad\\
{\textsuperscript{$2$} {Adobe Research} }\quad  
{\textsuperscript{$3$} {School of Interactive Computing, Georgia Institute of Technology} }\quad  \\
{\textsuperscript{$4$} {Department of Computer Science \& Engineering, Pennsylvania State University}}  \quad\\
\normalsize{\tt \{pranav.venkit, tfc5416, vkg5164, mukund, smr48, shomir\}@psu.edu}\\\normalsize{\tt hbiggs7@gatech.edu, koustavag@adobe.com}\\
}

\begin{document}
\maketitle

\begin{abstract}
We audit how hallucination in large language models (LLMs) is characterized in peer-reviewed literature, using a critical examination of 103 publications across NLP research. 
Through the examination of the literature, we identify a lack of agreement with the term `hallucination' in the field of NLP. Additionally, to compliment our audit, we conduct a survey with 171 practitioners from the field of NLP and AI to capture varying perspectives on hallucination. Our analysis calls for the necessity of explicit definitions and frameworks outlining hallucination within NLP, highlighting potential challenges, and our survey inputs provide a thematic understanding of the influence and ramifications of hallucination in society.
\end{abstract}

\section{Introduction}

\begin{figure}[h]
  \centering
  \includegraphics[scale = 0.38]{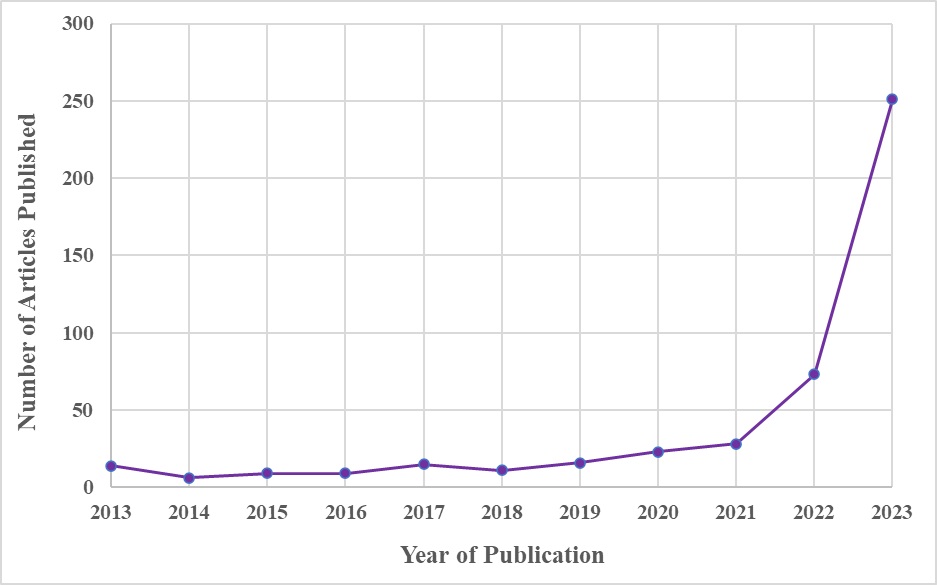}
  \caption{Articles published each year (from 2013 to 2023) in SCOPUS that contain the term `hallucination' AND (`NLP' OR `AI')  in the title, abstract, or keywords.}
  \vspace{-1.0 em}
  \label{fig:scopus}
\end{figure}

Recent advancements in Natural Language Processing (NLP) have expanded beyond traditional Machine Learning tools, evolving into sociotechnical systems that combine social and technical aspects to achieve specific goals \cite{gautam2024melting, narayanan2023towards}. They have now become integral in various domains such as health, policy-making, and entertainment, \cite{jin2022natural, werning2024generative} showcasing their significant impact on daily life.  
However, language models (LM) exhibit negative behaviours such as hallucination and biases \cite{bender2021dangers, gupta2024sociodemographic}. This has catalyzed a surge in research investigating the phenomenon of hallucinations in NLP \cite{ji2023survey}, reflected in the escalating number of peer-reviewed publications on the topic, as illustrated in Fig \ref{fig:scopus}, sourced from SCOPUS.

Within the NLP domain, various frameworks have emerged to define hallucination, primarily emphasizing its negative aspect. Hallucination here refers to the model's production of references to non-existent objects or statements, lacking supporting examples in the training data \cite{ji2023survey}. Despite the growing research on this topic, there is still a notable divide in our understanding, a lack of a unified framework, and a need for precise definitions \cite{filippova2020controlled}.

The necessity to understand this gap is accentuated by research demonstrating the societal impacts of hallucinations \cite{Dahl2024legal}. Hence, there is a growing need to explore how the field of NLP conceptualizes hallucination. In line with this imperative, the following questions guide this study:
\begin{itemize}
    \item\textbf{RQ1}: What are the definitions and frameworks used to explain hallucinations in NLP?
    \item\textbf{RQ2}: What is the current understanding of researchers about hallucinations, and how do they encounter them in their work?
\end{itemize}

To answer the RQ1, we first conduct an audit of the field of hallucinations in NLP by surveying 103 peer-reviewed articles\footnote{https://github.com/PranavNV/The-Thing-Called-Hallucination}. Subsequently, we conduct a survey to 171 researchers and academics in the field to gather their perspectives on this phenomenon, providing a novel contribution to the literature, addressing RQ2. By surveying NLP practitioners, the paper incorporates real-world perspectives, enriching the theoretical discussions with practical insights.
This audit therefore aims to broaden the communities perspective by presenting practical insights from researchers employing these methods in their work. We also propose an ethical framework to guide future efforts in comprehending and mitigating hallucinations in LLMs.

\section{Evolution of Hallucination in NLP}
The term `hallucination' has a long history in machine learning and has been used in various contexts prior to the LM era. Its earliest documented usage can be traced to the 2000s when \citet{baker-2000-7981} applied it in the context of image resolution enhancement, referring to the generation of new pixel values. Subsequently, "hallucination" has been frequently employed in computer vision research, including notable works such as \citet{hsu_face_hallucination} on face hallucination.


In the modern deep learning era, hallucination was used first by Andrej Karpathy in his blog focusing on Recurrent Neural Networks \cite{karpathy2015unreasonable}. He used the term within the context of LM by illustrating how an LSTM could generate non-existent URLs, effectively `hallucinating' data. The term then gained major traction with the launch of ChatGPT \cite{wu2023brief}, where it referred to inaccuracies and factual mistakes produced by models \cite{ji2023survey}. However, the field lacked a unified definition, leading to a spectrum of interpretations \cite{filippova-2020-controlled}. In one of the earlier works, \citet{maynez-etal-2020-faithfulness} divides term usage into intrinsic and extrinsic hallucination. Intrinsic hallucinations are consequences of synthesizing content using the information present in the input. Extrinsic hallucinations are model generations that ignore the source material altogether. 



However, there is a rise in discussion around terminology that reflects a deeper inquiry into the phenomena, with recent discourse advocating for `confabulation' \cite{millidge2023llms} or `fabrications' \cite{mcgowan2023chatgpt} as a more precise descriptor. This reflects the lack of consensus on the term and highlights the importance of looking at the use of hallucination with a more critical lens.

\section{Related Surveys on Hallucination}
We now provide an overview of several key surveys in the realm of NLP focusing on the topic of hallucination and why our work addresses a relevant gap in the field. Starting with \citet{ji2023survey}, this survey extensively delves into the advancements and challenges concerning hallucination in NLG, distinguishing between intrinsic and extrinsic frameworks of hallucination. Additionally, it sheds light on fundamental terms such as hallucination, faithfulness, and factuality, along with prevalent metrics for quantifying these phenomena. \citet{rawte2023survey} categorizes existing works within the domain of LMs, covering various aspects including methods for detecting hallucination, mitigation techniques, datasets used, and evaluation metrics. \citet{zhang2023siren} addresses the challenges of hallucination in LLMs by categorizing hallucinations into input-conflicting, context-conflicting, and fact-conflicting types, diverging from traditional viewpoints.

Furthermore, \citet{huang2023survey} redefines the taxonomy of hallucination into factuality and faithfulness, with additional subdivisions, and proposes mitigation strategies aligned with underlying causes. \citet{tonmoy2024comprehensive} offer a comprehensive overview of over thirty-two techniques developed to mitigate hallucination in LLMs and finally, \citet{rawte2023troubling} present a nuanced categorization of hallucination into six types, contributing to the ongoing discourse within the field.

While these surveys offer insights into the current state of hallucination research, they do not pay attention to critical examinations of the field's weaknesses arising from a lack of discourse in defining hallucination and challenges due to the same. This deficiency in discussion reflects the broader trend within the entire field. Therefore, our audit answers this gap by critically examining how we conceptualize hallucination. We aim to highlight the challenges stemming from these definitions and to further conduct a practitioner survey within the community to understand researchers' and developers' perspectives on this issue. 

\section{Critical Analysis of Hallucination in NLP Literature}

This section is dedicated to conducting an audit of hallucination research within NLP, aiming to uncover its applications and subsequently identify the strengths and weaknesses in current literature.

To accomplish this, we conducted an audit of works from the ACL anthology using specific keywords such as \textit{`hallucination', `NLP (OR) AI' AND `hallucinations', `fabrication',} and \textit{`confabulations'}. We surveyed papers released on and before \textit{April 19th, 2024}. From this search, a total of \textit{164 papers} were retrieved. After filtering out papers that were not directly related to hallucination research or those that merely mentioned the term without substantial focus on the topic, we arrived at a corpus of \textbf{103 papers}. This corpus forms the basis for our audit and analysis of hallucination research, specifically within the NLP domain.

\subsection{Conceptualization of Hallucination}

We performed an iterative thematic analysis \cite{vaismoradi2013content} to uncover the various applications of hallucination research in NLP. To ensure accuracy and prevent misclassification, this recursive process was employed. This resulted in the identification of seven distinct fields that address research on hallucination (as shown in Table. \ref{table:categories}).

\begin{table}[]
\centering
\small
\begin{tabular}{c c}
\hline
\textbf{NLP Tasks} & \textbf{Frequency} \\ \hline
Conversational AI & 38 \\
Abstractive Summarization & 16 \\
Data-to-Text Generation & 14 \\
Machine Translation & 12 \\
Image-Video Captioning & 8 \\
Data Augmentation &  8\\
Miscellaneous &  7\\ \hline
\end{tabular}
\vspace{-0.5 em}
\caption{Frequency of papers reviewed for each thematically grouped NLP tasks.}
\vspace{-1.2 em}
\label{table:categories}
\end{table}

This taxonomy affords insights into the pervasive nature of hallucination in NLP. Notably, it reveals that hallucination transcends beyond text generation, extending its conceptualization to encompass broader domains such as \textit{Image-Video Captioning, Data Augmentation,} and \textit{Data-to-Text Generation} tasks. This depicts the significance of hallucination both within and beyond the realm of NLP. Moreover, our classification framework provides us with a faceted analysis of how each of these tasks defines the concept of hallucination.

Using thematic categorization, we come across definite attributes across the definitions of hallucination. One set of attributes elucidated how hallucinations are associated with the style/language generated by the model: \textbf{Fluency, Plausibility}, and \textbf{Confidence}.  
The next set of attributes falls under the effects of hallucinations: \textbf{Intrinsic, Extrinsic, Unfaithfulness} and \textbf{Nonsensical}. The definition of each of these attributes is elaborated in Table \ref{table:definitions}. 

In each paper analyzed within the survey scope, hallucination is defined based on a combination of the set of attributes identified. Our survey revealed \textbf{31 unique frameworks} for conceptualizing hallucination, illustrating the diverse approaches and perspectives used. This diversity underscores the ambiguity in the term's usage.

To illustrate this phenomenon, we present some examples showcasing the diverse approaches commonly observed in the literature:

\textit{\say{Hallucination refers to the phenomenon where the model generates false information not supported by the input.} - \cite{xiao2021hallucination}}

\textit{\say{LLMs often exhibit a tendency to produce exceedingly confident, yet erroneous, assertions commonly referred to as hallucinations.} - \cite{zhang-etal-2023-sac3}}

\textit{\say{Models generate plausible-sounding but unfaithful or nonsensical information called hallucinations} - \cite{ji2023towards}}

Hence, within NLP, a notable deficiency persists in grasping coherent characteristics of hallucination. This shortfall underscores the risk of potential misappropriation of the term when employed in divergent contexts. An extensive analysis of hallucination for each of the mentioned NLP tasks and its definition is illustrated in the \textit{Appendix \ref{appendix-NLPtask}}.

\begin{table*}[]
\footnotesize
\centering
\begin{tabular}{|c|c|}
\hline
\textbf{Attributes} & \textbf{Definition} \\ \hline
Fluency &  \begin{tabular}[c]{@{}c@{}}The syntactic incorrectness and semantic errors of the sentence generated. \end{tabular}\\
\hline
Plausibility & \begin{tabular}[c]{@{}c@{}}The degree to which the generated text\\ appears factually incorrect or unbelievable within the given context.\end{tabular}\\
\hline
Confidence & \begin{tabular}[c]{@{}c@{}}The absence of modifiers or qualifiers that express \\uncertainty in the generated text, presenting the output with a sense of assuredness.\end{tabular} \\
\hline
Intrinsic & \begin{tabular}[c]{@{}c@{}}The generated output that contradicts the source content or the input provided\end{tabular} \\
\hline
Extrinsic & \begin{tabular}[c]{@{}c@{}}The generated output that cannot be verified from the source content or the input provided.\end{tabular} \\
\hline
Non-factual &  \begin{tabular}[c]{@{}c@{}}Inconsistent with facts in the real world, leading to the generation of non-factual\\ content in accordance to the established real-world knowledge.\end{tabular}\\
\hline
Unfaithfulness & \begin{tabular}[c]{@{}c@{}}Inconsistent to the input prompt or context, creating deviations\\ or inconsistencies that would diverge from the intended meaning or message.\end{tabular} \\
\hline
Nonsensical & \begin{tabular}[c]{@{}c@{}}Lack of logical meaning or coherence within a given context as well as the readability of the text.\end{tabular} \\
\hline
\end{tabular}
\vspace{-0.5 em}
\caption{The attributes that appear in the definitions of hallucination.} 
\vspace{-1.2 em}

\label{table:definitions}

\end{table*}

\subsection{Audit of Frameworks}


We now scrutinize the dominant frameworks employed in defining hallucination while also assessing the extent to which these models accurately capture the phenomenon. 
We start by looking at \textit{how many of the selected works explicitly define hallucination.} Out of the \textbf{103} papers, just \textbf{44} (42.7\%) provide a definition of the term, leaving the majority—\textbf{59 papers or 57.3\%}—either altogether omitting their understanding of hallucination in the context of their research or providing no definition or a framework. This lack of transparency is not only concerning but also underscores the need for clarity, especially given the varied interpretations of hallucination across different research domains.

Taking our scrutiny a step further, we investigate \textit{whether the works defining hallucination reference and acknowledge preexisting frameworks}. It emerges that only \textbf{29 papers or 27\%}  of the selected works explicitly acknowledge and adhere to established frameworks of hallucination, while the remainder \textbf{73\%} either loosely define the term or devise new definitions tailored to their specific research scope. This trend within the field shows a lack of consensus on the conceptualization of hallucination, leading to disparate interpretations and a shortage of discourse on the subject.

We also audit the sociotechnical nature of the definitions of hallucination in NLP. Hallucination (elucidated in Appendix \ref{appendix-social}) inherently contains social dimensions, creating varied perspectives across different social contexts. 
Moreover, given the evolution of LMs into social spaces, adopting a sociotechnical approach becomes necessary, given that the term `hallucination' is inherently a \textit{shared vocabulary} within these domains. Unfortunately, out of the 103 works examined, \textbf{only 3} acknowledge the this nature of hallucination, with \textbf{none} utilizing this framework to inform their approach. This underscores a need for research to explore the sociotechnical dimensions inherent in hallucination, showcasing the limited depth of understanding within the ML and NLP communities.

\subsection{Audit of Metrics}

In the analysis of the 103 papers, we observed that \textbf{87 of these works} dedicate efforts to measuring `hallucination.' This observation depicts the prevailing trend within NLP, emphasizing the significance of quantifying the concept of hallucination across diverse research efforts. Building upon prior studies such as \citet{ji2023survey}, our analysis categorizes the common approaches in NLP for quantifying hallucination into four major themes: 
\textit{Statistical Metrics, Data-driven Metrics, Human Evaluation,} and \textit{Mixed Methodologies}.


\begin{figure}[h]
  \centering
  \includegraphics[scale = 0.16]{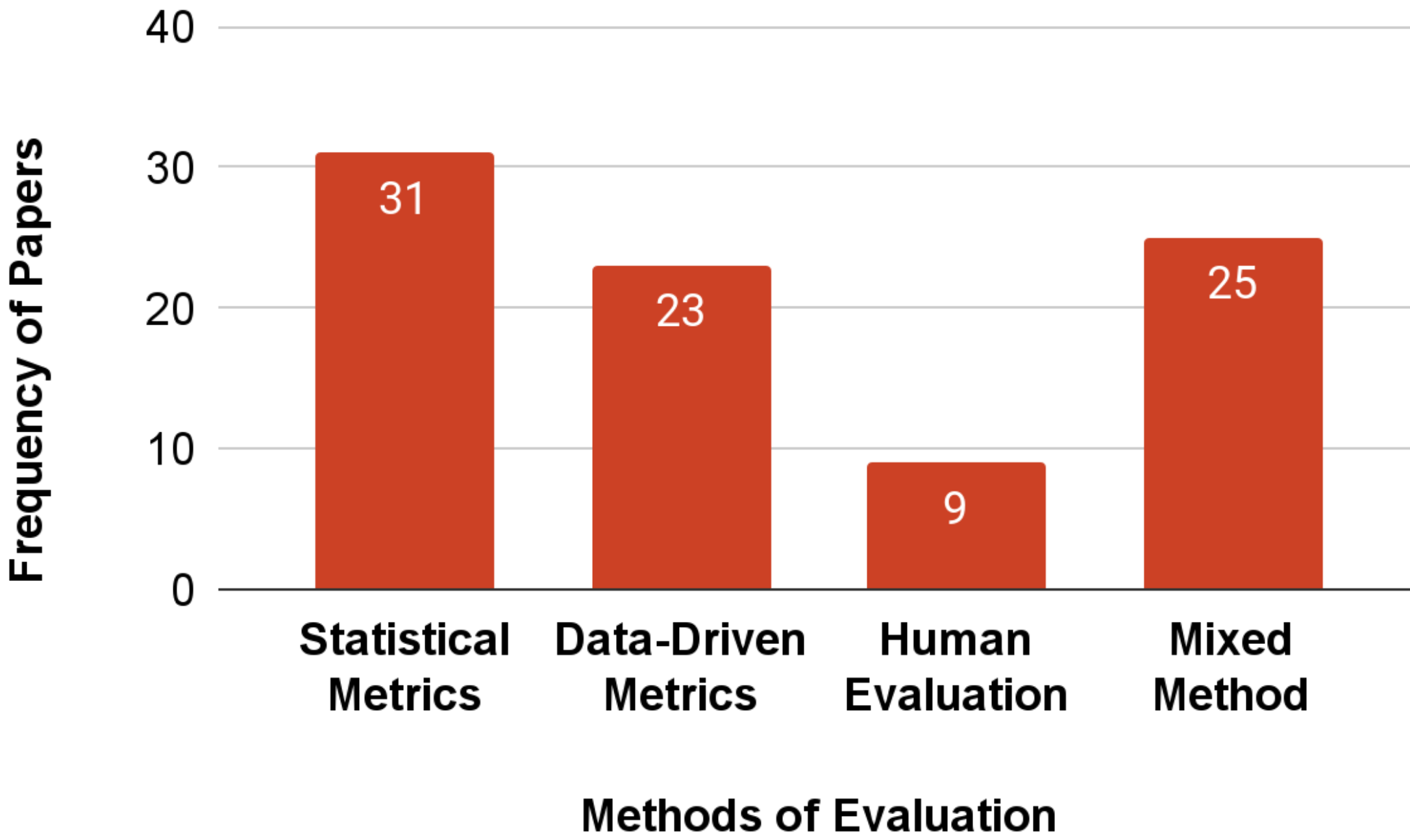}
  \vspace{-0.5 em}
  \caption{Hallucination evaluation metrics used in NLP.}
  \vspace{-1.2 em}
  \label{fig:metrics}
\end{figure}

\textbf{Statistical metrics} 
calculates a hallucination score based on the degree of mismatch, with higher discrepancies indicating lower accuracy, factuality or faithfulness and hence, higher hallucination \cite{ji2023survey}. Statistical scores such as BLUE, ROUGE, and Error Rate metrics are commonly used in this approach. Our findings reveal that \textbf{35.2\%} of the works that quantify hallucination opt for statistical metrics, employing \textbf{25 distinct metrics} (e.g., BERTScore, F1, Perplexity, Cosine Similarity) developed for this purpose. 
This variability underscores the lack of a standardized approach.

\textbf{Data-driven metrics} utilizes curated datasets or neural models to gauge hallucination in generated text. This methodology, accounting for curated knowledge/content mismatches, is adopted by \textbf{26.1\%} of the works, resulting in the development of \textbf{18 distinct} datasets or models tailored for hallucination measurement, such as CHAIR (Caption Hallucination Assessment with Image Relevance) and SelfCheckGPT \cite{manakul-etal-2023-selfcheckgpt}.

\textbf{Human evaluation} offers a complementary perspective by employing human annotators to assess hallucination levels, compensating for apparent errors in automated indicators \cite{ji2023survey}. This approach, used by \textbf{10.2\%} of the works, encompasses scoring and comparison methods, where annotators rate hallucination levels or compare output texts with baselines or ground-truth references. Notably, one outlier paper introduced an innovative approach utilizing eye tracking for hallucination detection in NLP tasks \cite{maharaj-etal-2023-eyes}.

\textbf{Mixed method} approach is deployed by \textbf{28.4\%} of the works, combining human evaluation with statistical metrics to offer a holistic perspective on hallucination quantification. This trend reflects a concerted effort within the research community to address the limitations of individual methodologies and provide insights into the presence and nature of hallucination in generated texts.

The metrics audit reveals significant knowledge gaps and challenges across various approaches. Notably, established research highlights areas for improvement in standard methods for measuring hallucination. For instance, methodologies like CHAIR and metrics such as ROUGE scores exhibit instability in measuring hallucination due to the need for complex human-crafted parsing rules for exact matching, rendering them susceptible to errors \cite{li-etal-2023-evaluating}. Criticisms also extend to human evaluation methods, which are prone to inaccuracies in gauging hallucination within these models \cite{smith-etal-2022-human}.

Beyond methodological criticisms, our audit uncovers a trend of employing numerous distinct metrics and approaches within these frameworks to categorize hallucinations. Over time, this has led to a diverse set of parameters for measuring hallucination, with a general lack of consensus on a standardized measurement approach. 
This issue further highlights the absence of a unified method, especially as these models have now shifted to become a sociotechnical solution \cite{bender2021dangers}.

\section{Practitioner Survey of Hallucination}

In this section, adopting a `community-centric approach' \cite{narayanan2023towards}, we conduct a survey to gain insights into researchers' perceptions of hallucinations in NLP to complement our theoretical discussions with practical real-world perspectives.
The primary goal is to demonstrate how researchers and practitioners within the field perceive the concept of `hallucination' and to expand our findings beyond the limitations of existing literature where real-world perceptions from the researchers are missing \cite{huang2023survey, zhang2023siren, ji2023survey}.
This motivates us to gather real-world perspectives from individuals actively engaged in NLP and AI research.

\subsection{Survey Recruitment and Data Collection}
For our survey, we employed a multi-faceted approach to reach a diverse population of respondents. We utilized direct emails, direct messages, and social media platforms such as LinkedIn and Twitter to distribute the survey. Our target audience included graduate students and professors from academic backgrounds as well as individuals from the industry who work in NLP, aiming to capture a wide range of perspectives on hallucinations.

To ensure a comprehensive view, we specifically targeted researchers familiar with AI and ML, primarily from disciplines such as computer science and information science. However, we also welcomed participants from other domains to explore their perceptions of whether they had the literature understanding of the concept of hallucination as they are also extensively using LLM models. The survey was examined and approved by the Institutional Review Board (IRB) for ethical practices.

We additionally employed a systematic approach by randomly selecting 15 universities from the top 100 in the USA as per the 2023 US News and World Report rankings \cite{us_ranking}, to then reach out to potential participants. Prior works \cite{chakravorti2023prototype} have previously employed this process to identify high-quality participants.
We received a total of 223 responses, out of which 171 were complete and usable for analysis.


\subsection{Survey Structure}

The survey employed a combination of 14 open-ended and close-ended questions. The survey has been built based on the previous survey design techniques \cite{rosen2013media, baker2016reproducibility, van2023ai, chakravorti2024reproducibility}. Open-ended questions and free-response text boxes allow us to gather rich opinions from participants. This approach integrates all our findings, providing a broader and deeper understanding of the response. For the analysis of open-ended questions, we utilized thematic analysis, drawing from established methodologies outlined in \citet{blandford2016qualitative, terry2017thematic}. 
The close-ended questions were analyzed using descriptive statistics to summarize and analyze the numerical data obtained from respondents. 
Throughout the analysis process, the research team made collective decisions regarding the retention, removal, or reorganization of themes derived from open-ended responses. All the survey questions have been provided in \textit{Appendix \ref{appendix_questions}}.

\subsection{Survey Findings}
We now summarize insights from our responses to explore various perspectives on hallucinations in LLMs, including perceptions, weaknesses, and preferences. The breakdown of responses indicates that 76.54\% of participants were from academia, 20.98\% from the industry, and 2.47\% both.

Participants were also asked about their research area's direct relation to AI and NLP. The analysis revealed that more than 68.52\% of researchers indicated that their work is directly related to NLP, while the remaining respondents either exhibited familiarity with or indirectly incorporated NLP and AI methodologies in their work. This highlights the substantial involvement of AI experts and practitioners within the survey.

\subsubsection{Familiarity with Hallucination}
\begin{figure}[t]
  \centering
  \includegraphics[scale = 0.21]{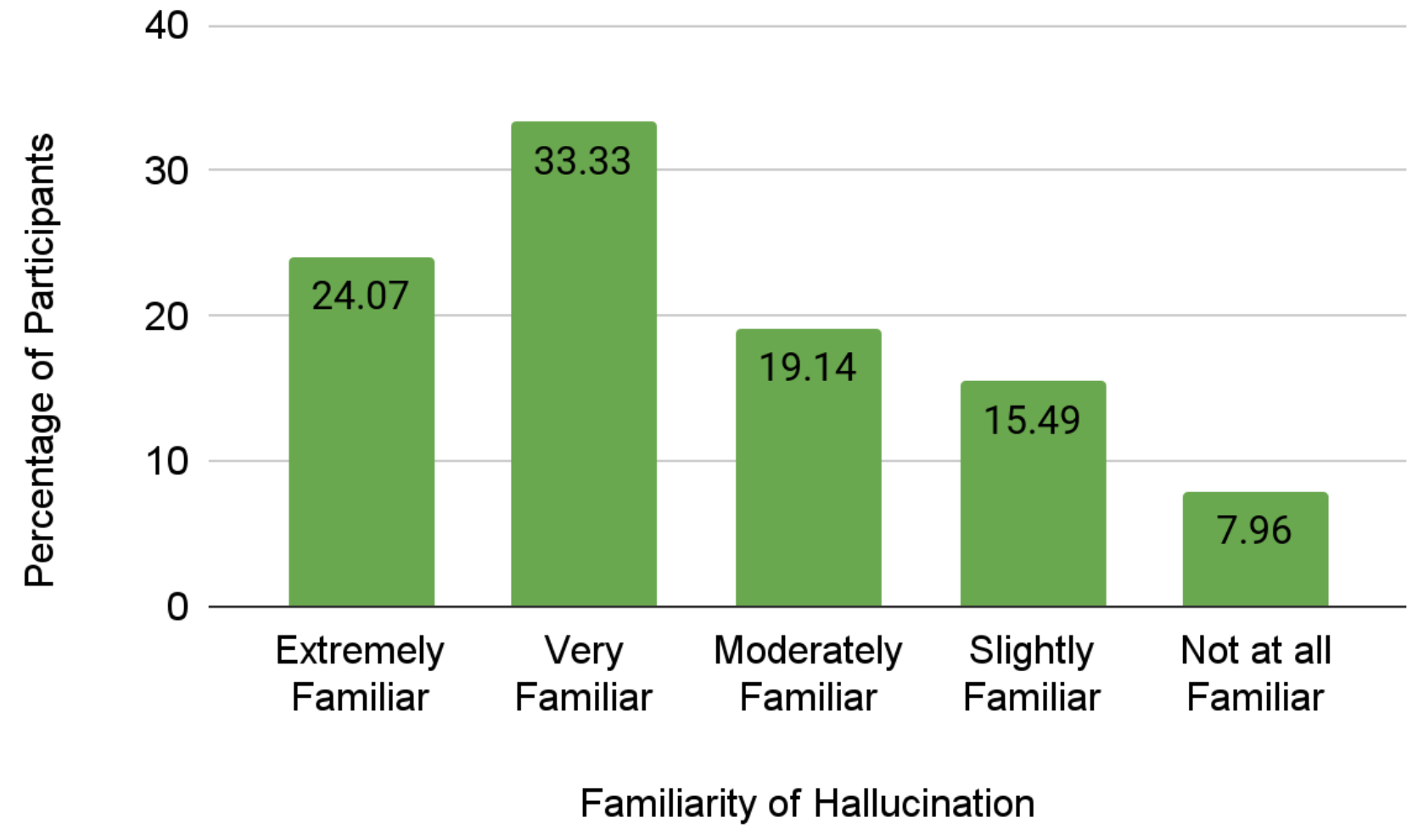}
  \vspace{-0.5em}
  \caption{Respondents familiarity with `Hallucination'}
  \vspace{-1.0 em}
  \label{fig:Familiarity}
\end{figure}

The survey included the question on participants' familiarity with the concept of `hallucinations' in AI-generated text, measured on a 5-point Likert scale. The analysis revealed that 24.07\% of researchers reported being extremely familiar with the concept, while 33.33\% indicated being very familiar with it (Figure \ref{fig:Familiarity}). 
Participants who indicated not being familiar with the term `hallucination' (7.96\%) also demonstrated implicit concerns with this phenomenon by highlighting issues such as generating incorrect responses and crafting stories autonomously. This demonstrates the widespread impact of the phenomenon within the community.

\begin{figure}[b]
  \centering
  \vspace{-0.5em}
  \includegraphics[scale = 0.22]{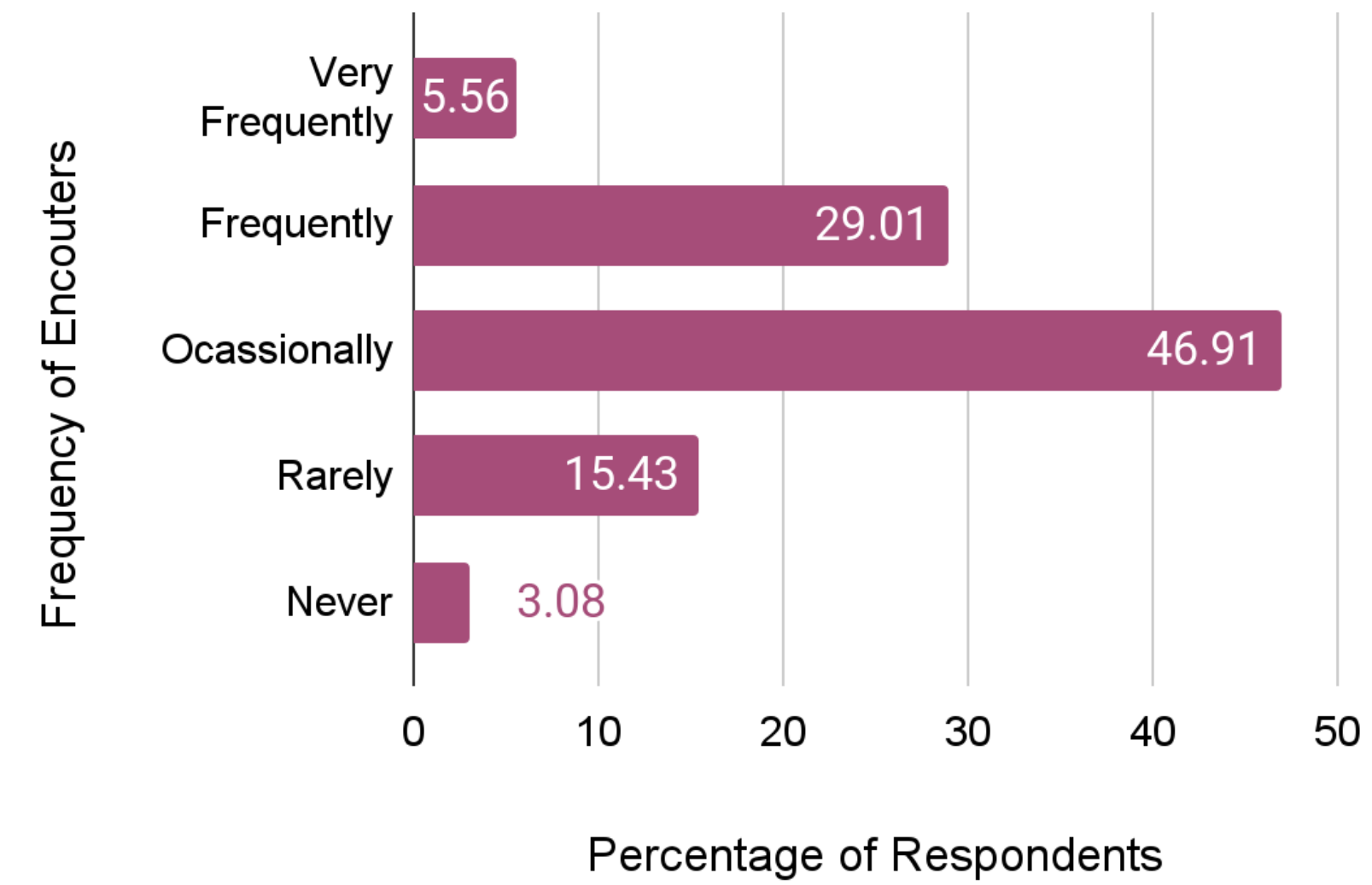}
  \vspace{-0.5em}
  \caption{Frequency of encountering `Hallucination'}
  \label{fig:Frequency}
\end{figure}

\subsubsection{Hallucination Frequency}
The survey included a question regarding the frequency of encountering `hallucinated' content, defined as content that is factually incorrect or unrelated to the input, assessed on a 5-point Likert scale ranging from `Never' to `Very frequently' (Figure \ref{fig:Frequency}). The analysis revealed that 46.91\% of respondents reported encountering hallucinated content occasionally, while 29.01\% indicated experiencing it frequently. The results suggest that a substantial portion of practitioners encounter instances of hallucinated content in AI-generated outputs, indicating a prevalent issue in generative NLP models.

\subsubsection{Perceptions of Hallucination}
The survey findings revealed that \textbf{more than 92\%} of respondents perceive hallucination as a weakness of LLMs. Subsequently, participants were asked to provide their own definitions of `hallucination' in generative AI models through an open-ended question. To analyze these responses, we applied thematic categorization based on attributes generated from the literature audit (Table \ref{table:definitions}).

The thematic categorization revealed that the majority of respondents categorized hallucination as pertaining to the\textit{ factuality and faithfulness of input, with relatively lesser emphasis on the extrinsic and intrinsic nature of hallucination} concerning the input. This trend reflects a common perception of how hallucination is understood within the context of larger-scope generative AI models.

Moreover, the analysis identified \textbf{12 distinct frameworks} regarding how hallucination is defined by respondents. For example:

\textit{\say{Response that appears syntactically and semantically believable, but is not based on actual fact}}---Academic Researcher, NLP

\textit{\say{When the model confidently states something that is not true}}---Academic Researcher, AI


The diversity of viewpoints underscores the inconsistency within the field regarding the conceptualization and understanding of hallucination in the context of generative AI models.

\subsubsection{Alternative Terms for Hallucination}
The survey included a question asking participants if they prefer an alternate term to describe the phenomenon of `hallucination' in AI-generated content and to provide an explanation if they do. The analysis revealed that 54.32\% of respondents preferred the term hallucination or had no other term to provide. However, among the remaining responses,\textbf{ 40.46\%} of participants mentioned \textbf{`Fabrication'} as a better term to describe the phenomenon.

This indicates that while the majority of respondents did not propose an alternative term, a notable proportion sees fabrications as a more suitable descriptor for the phenomenon of hallucination in AI-generated content. For example, 

\textit{\say{Fabrication makes more sense. Hallucination makes it feel like AI is human and has the same sensory perceptions that could lead to hallucinations.}}---Academic Researcher, AI \& Education

It's interesting to note that a few researchers also prefer to use the term \textbf{`Confabulations'} instead of `hallucinations' when referring to AI-generated content. Their rationale likely stems from the nuanced difference in meaning between the two terms. While hallucinations generally convey the idea of perceiving something that is not based on reality or fact, confabulations specifically refer to the creation of false memories or information without the intention to deceive.

By opting for the term `Confabulations,' researchers may be emphasizing the unintentional nature of the inaccuracies or false information generated by AI models, as opposed to implying deliberate deceit. For example,

\textit{\say{I think confabulation works better because it means creating a false memory without deceit. Fabrication gives the idea that it is intentional, which in the case of generative AI models, it is not.}}---Academic Researcher, AI \& HCI

It's also insightful to see that respondents proposed various alternative terms to describe the phenomenon of hallucination in AI-generated content such as \textit{incorrect information/misinformation, Non-factual information, Cognitive gap, hyper-generalization, Overconfidence,} and \textit{Randomness}. These alternatives highlight different aspects and nuances of the inaccuracies or distortions present in the generated content. Participants also mentioned how they prefer multiple terms based on the application in which they are used.

\textit{\say{As I mentioned there are different types of hallucinations. For instruction and context hallucinations, I would refer to them as inconsistency instead. For factually incorrect hallucinations, the word hallucination is fine.}}---Academia, NLP

\subsubsection{Creativity and Positive Applications}
Not all researchers view hallucinations in AI-generated content through a negative lens. While the majority may associate hallucinations with inaccuracies or distortions, a notable minority \textbf{($\sim$12\%} in our survey) provided insights into how they believe hallucinations in these models can be correlated with creativity rather than negatively impacted behaviors. In fields such as story narration and image generation, researchers often value the creative behaviors exhibited by AI models. Hallucinations, when viewed in this context, may be seen as manifestations of the model's ability to think outside the box, generate novel ideas, and explore unconventional possibilities. These creative outputs can inspire new approaches to storytelling, art, and problem-solving, contributing to innovation and artistic expression. For example:

\textit{\say{Hallucinations are just what is needed for models to be creative. In truth, unless AI text-generators are factually grounded with external knowledge for a specific field, they are just story generators which aim to be creative, hence``hallucinate."}}---National Lab Researcher, NLP

Further supplementary analysis and quotes on the various external perspectives and the societal ramifications of hallucination, obtained through the survey, is examined in the \textit{Appendix \ref{appendix-survey}}.

\section{Challenges and Recommendations}

Based on our audit and survey analysis, we outline the key weaknesses encountered in hallucination within NLP and potential recommendations motivated by the weaknesses. We utilize a community-centric approach to define the primary weaknesses of the field currently and a path forward.

\subsection{Challenges}
The primary challenges we identify thematically and aim to elucidate are as follows:

\textbf{Wide range of vague and absent definitions:} The literature and the practitioner's survey show diverse and conflicting frameworks, often lacking clarity or omitting explicit definitions for hallucination and how it is perceived in various fields of NLP and language generation. Ambiguity arises from the use of terms like `confabulations,' ` fabrications,' `misinformation,' and `hallucinations' interchangeably, without clear definitions in the context of hallucinations.
 
\textbf{Lack of standardization in measurement:} The absence of standardized metrics to quantify hallucination results in the use of multiple scales and categorizations. This makes it challenging to compare and interpret results across different models and studies, leading to a proliferation of diverse approaches for evaluating hallucinations.

\textbf{Limited awareness of hallucination in a sociotechnical context:} Hallucination research often lacks the understanding of how the concept of hallucination is conceptualized beyond its technical purview. When such analysis is employed in sociotechnical systems like healthcare and policy making \cite{Dahl2024legal, pal-etal-2023-med}, it is important to define the socially relevant framework of hallucination as well. There is no motivation shown to understand the non-technical considerations of hallucination.



\textbf{Multiple sentiment towards hallucination:}
The perception of hallucination in generative AI varies depending on the context. For instance, it is often positively regarded as creativity in image generation, whereas in text generation, it is viewed negatively as errors or mistakes. Consequently, future research efforts should aim to better address this disparity to develop a more nuanced framework for understanding hallucination.

\textbf{Lack of standardized nomenclature:} Both our literature audit and practitioner survey revealed that the term `hallucination' is inadequate to fully capture the behavior exhibited by NLG models. There is a need for further investigation into which terms are more appropriate and why they are necessary. For instance, terms like `confabulation,' `fabrications,' and `misinformations' are increasingly being used to describe the same phenomenon. A more precise understanding is required to distinguish between these terms and how they are utilized in various fields within NLP.

\textbf{User trust and reliability:} Our survey findings suggest that users may hesitate to fully utilize LLM capabilities due to concerns about bias and hallucination despite recognizing the potential advantages these models offer. Therefore, there is a need to focus efforts on understanding the human interaction aspect concerning hallucination in NLP and language generation.


Addressing these issues requires careful consideration of the categorization approach, integration of contextual information, and, efforts towards robust evaluation methodologies in hallucinations.

\subsection{Recommendations}
Expanding on audits like \citet{blodgett-etal-2020-language} \& \citet{venkit2023sentiment}, we examine strategies for NLP practitioners studying `hallucination' to overcome these challenges. We propose two overarching themes with four associated recommendations.

\paragraph{Author-Centric Recommendation.}

These recommendations prioritize actionable steps for both the author and developers, emphasizing transparent and accountable development in conceptualizing hallucinations.

\textbf{[R1]} Ensure explicit documentation of the hallucination framework and analysis methodology employed during the development of NLP models. Provide guidelines that outline the expected measurements and quantifications for the model to enhance interpretability and applicability.

\textbf{[R2]} Explicitly state the use cases and user profiles intended to interact with the NLP system. By considering the specific applications and targeted users, it is easier to construct the required framework of hallucination that is sensitive to the community in consideration. Raise awareness about potential ramifications introduced by NLP models, emphasizing the importance of fairness and equity. 

\paragraph{Community-Centric Recommendation.}
These recommendations prioritize actionable steps for the research community to enhance frameworks and understanding related to hallucinations.

\textbf{[R3]} Develop clear and standardized definitions for terms such as `confabulations,' `fabrications,' `misinformation,' and `hallucinations' within the context of NLP. Establish frameworks that provide clarity and consistency in understanding these concepts, particularly regarding hallucinations. This requirement is crucial due to the widespread misunderstanding of hallucination and the misnomers that have arisen as research progresses.

\textbf{[R4]} Promote the creation of methods that offer visibility into the model's decision-making process, enabling users to comprehend how hallucinations or fabrications can occur within the system, thus fostering trust in its use. Facilitating research discussions for transparency through workshops and conferences is one approach to achieving this goal.


\section{Conclusion}
Our work delves into the conceptualization of hallucination within the scope of NLP. Our approach involved two key methodologies: first, an exhaustive audit of 103 peer-reviewed papers in the NLP domain, and second, a practitioner survey of 171 researchers to complement our first study with real-world practical perception and understanding of hallucination as a unique contribution. Through this analysis, we have gained insights into how the NLP community conceptualizes and defines hallucination, showcasing a lack of discourse and agreement. Additionally, our thematic and community-based approach highlights potential weaknesses within the field, particularly in addressing misrepresentations and inaccurate characterizations associated with hallucination, paving way for better advancement in language generation. 

\section{Limitations}

Our study encompasses a selection of 103 papers, incorporating works from primarily the ACL Anthology. While our intention was not to provide an exhaustive collection of all published works on hallucination, we aimed to include diverse sources within NLP that cover various aspects of the field. Our intent was to curate peer-reviewed literature commonly found in the NLP domain, encompassing models, applications, survey papers, and frameworks.
We, therefore, did not scope the utility of hallucination and its impact beyond NLP to other fields of research, such as Computer Vision. Regarding the creation of the challenges and recommendations, it is important to note that the themes presented are not meant to be exhaustive but rather serve as a foundational framework to spark additional inquiries and foster further engagement. 

Our survey was designed to capture the viewpoints of researchers and practitioners in the AI and ML field, potentially limiting various experiences. As such, our analysis is centered on this perspective. While we did gather additional insights from participants outside this field, our focus was not comprehensive in that regard. Our future work intends to explore the public's perspective on hallucination.

\section{Ethics Statement}

We are aware of the ethical considerations involved in our research and have taken measures to ensure responsible practices throughout the study.

Data Publication: All the papers used in our research are listed in the Appendix. However, we recognize the importance of transparency and accountability. Therefore, we publish the complete collection of papers along with our qualitative classification and annotation, allowing for public scrutiny and examination \footnote{https://github.com/PranavNV/The-Thing-Called-Hallucination}.

Mitigating Qualitative Bias: We acknowledge the potential for bias when performing qualitative coding of the papers regarding their applications. To address this concern, we ensured that at least three different individuals independently reviewed and verified the coding to minimize the possibility of misclassification. Additionally, we followed the same approach to verify the presence of various definitions in each paper, enhancing the reliability and validity of our analysis.
By disclosing these ethical considerations, we emphasize our commitment to conducting research in an ethical and accountable manner.

\bibliography{tacl2021}

\begin{thebibliography}{125}
\expandafter\ifx\csname natexlab\endcsname\relax\def\natexlab#1{#1}\fi

\bibitem[{Akani et~al.(2023)Akani, Favre, Bechet, and Gemignani}]{akani2023reducing}
Eunice Akani, Benoit Favre, Frederic Bechet, and Romain Gemignani. 2023.
\newblock Reducing named entity hallucination risk to ensure faithful summary generation.
\newblock In \emph{Proceedings of the 16th International Natural Language Generation Conference}, pages 437--442.

\bibitem[{Anastasopoulos and Neubig(2019)}]{anastasopoulos2019pushing}
Antonios Anastasopoulos and Graham Neubig. 2019.
\newblock Pushing the limits of low-resource morphological inflection.
\newblock \emph{arXiv preprint arXiv:1908.05838}.

\bibitem[{Baker(2016)}]{baker2016reproducibility}
Monya Baker. 2016.
\newblock Reproducibility crisis.
\newblock \emph{nature}, 533(26):353--66.

\bibitem[{Baker and Kanade(2000)}]{baker-2000-7981}
Simon Baker and Takeo Kanade. 2000.
\newblock Hallucinating faces.
\newblock In \emph{Proceedings of 4th IEEE International Conference on Automatic Face and Gesture Recognition (FG '00)}, pages 83 -- 88.

\bibitem[{Bender et~al.(2021)Bender, Gebru, McMillan-Major, and Shmitchell}]{bender2021dangers}
Emily~M Bender, Timnit Gebru, Angelina McMillan-Major, and Shmargaret Shmitchell. 2021.
\newblock On the dangers of stochastic parrots: Can language models be too big?
\newblock In \emph{Proceedings of the 2021 ACM conference on fairness, accountability, and transparency}, pages 610--623.

\bibitem[{Berbatova and Salambashev(2023)}]{berbatova-salambashev-2023-evaluating}
Melania Berbatova and Yoan Salambashev. 2023.
\newblock \href {https://aclanthology.org/2023.ranlp-stud.6} {Evaluating hallucinations in large language models for {B}ulgarian language}.
\newblock In \emph{Proceedings of the 8th Student Research Workshop associated with the International Conference Recent Advances in Natural Language Processing}, pages 55--63, Varna, Bulgaria. INCOMA Ltd., Shoumen, Bulgaria.

\bibitem[{Blandford et~al.(2016)Blandford, Furniss, and Makri}]{blandford2016qualitative}
Ann Blandford, Dominic Furniss, and Stephann Makri. 2016.
\newblock \emph{Qualitative HCI research: Going behind the scenes}.
\newblock Morgan \& Claypool Publishers.

\bibitem[{Blodgett et~al.(2020)Blodgett, Barocas, Daum{\'e}~III, and Wallach}]{blodgett-etal-2020-language}
Su~Lin Blodgett, Solon Barocas, Hal Daum{\'e}~III, and Hanna Wallach. 2020.
\newblock \href {https://doi.org/10.18653/v1/2020.acl-main.485} {Language (technology) is power: A critical survey of {``}bias{''} in {NLP}}.
\newblock In \emph{Proceedings of the 58th Annual Meeting of the Association for Computational Linguistics}, pages 5454--5476, Online. Association for Computational Linguistics.

\bibitem[{Boeving(2020)}]{boeving2020transpersonal}
Nicholas~Grant Boeving. 2020.
\newblock Transpersonal psychology.
\newblock In \emph{Encyclopedia of Psychology and Religion}, pages 2392--2394. Springer.

\bibitem[{Bouyamourn(2023)}]{bouyamourn-2023-llms}
Adam Bouyamourn. 2023.
\newblock \href {https://doi.org/10.18653/v1/2023.emnlp-main.192} {Why {LLM}s hallucinate, and how to get (evidential) closure: Perceptual, intensional, and extensional learning for faithful natural language generation}.
\newblock In \emph{Proceedings of the 2023 Conference on Empirical Methods in Natural Language Processing}, pages 3181--3193, Singapore. Association for Computational Linguistics.

\bibitem[{Cao et~al.(2021)Cao, Dong, and Cheung}]{cao2021hallucinated}
Meng Cao, Yue Dong, and Jackie Chi~Kit Cheung. 2021.
\newblock Hallucinated but factual! inspecting the factuality of hallucinations in abstractive summarization.
\newblock \emph{arXiv preprint arXiv:2109.09784}.

\bibitem[{Chakravorti et~al.(2023)Chakravorti, Fraleigh, Fritton, McLaughlin, Singh, Griffin, Kwasnica, Pennock, Giles, and Rajtmajer}]{chakravorti2023prototype}
Tatiana Chakravorti, Robert Fraleigh, Timothy Fritton, Michael McLaughlin, Vaibhav Singh, Christopher Griffin, Anthony~Mark Kwasnica, David Pennock, C~Lee Giles, and Sarah Rajtmajer. 2023.
\newblock A prototype hybrid prediction market for estimating replicability of published work.
\newblock In \emph{2nd International Conference on Hybrid Human-Artificial Intelligence, HHAI 2023}, pages 300--309. IOS Press BV.

\bibitem[{Chakravorti et~al.(2024)Chakravorti, Koneru, and Rajtmajer}]{chakravorti2024reproducibility}
Tatiana Chakravorti, Sai~Dileep Koneru, and Sarah Rajtmajer. 2024.
\newblock Reproducibility, replicability, and transparency in research: What 430 professors think in universities across the usa and india.
\newblock \emph{arXiv preprint arXiv:2402.08796}.

\bibitem[{Chen et~al.(2023{\natexlab{a}})Chen, Wu, and Li}]{chen2023exploring}
Qinyu Chen, Wenhao Wu, and Sujian Li. 2023{\natexlab{a}}.
\newblock Exploring in-context learning for knowledge grounded dialog generation.
\newblock In \emph{Findings of the Association for Computational Linguistics: EMNLP 2023}, pages 10071--10081.

\bibitem[{Chen et~al.(2021)Chen, Zhang, Sone, and Roth}]{chen-etal-2021-improving}
Sihao Chen, Fan Zhang, Kazoo Sone, and Dan Roth. 2021.
\newblock \href {https://doi.org/10.18653/v1/2021.naacl-main.475} {Improving faithfulness in abstractive summarization with contrast candidate generation and selection}.
\newblock In \emph{Proceedings of the 2021 Conference of the North American Chapter of the Association for Computational Linguistics: Human Language Technologies}, pages 5935--5941, Online. Association for Computational Linguistics.

\bibitem[{Chen et~al.(2023{\natexlab{b}})Chen, Wu, Chen, and Chen}]{chen2023fidelity}
Wei-Lin Chen, Cheng-Kuang Wu, Hsin-Hsi Chen, and Chung-Chi Chen. 2023{\natexlab{b}}.
\newblock Fidelity-enriched contrastive search: Reconciling the faithfulness-diversity trade-off in text generation.
\newblock \emph{arXiv preprint arXiv:2310.14981}.

\bibitem[{Choubey et~al.(2023)Choubey, Fabbri, Vig, Wu, Liu, and Rajani}]{choubey-etal-2023-cape}
Prafulla~Kumar Choubey, Alex Fabbri, Jesse Vig, Chien-Sheng Wu, Wenhao Liu, and Nazneen Rajani. 2023.
\newblock \href {https://doi.org/10.18653/v1/2023.findings-acl.685} {{C}a{PE}: Contrastive parameter ensembling for reducing hallucination in abstractive summarization}.
\newblock In \emph{Findings of the Association for Computational Linguistics: ACL 2023}, pages 10755--10773, Toronto, Canada. Association for Computational Linguistics.

\bibitem[{Cirik et~al.(2022)Cirik, Morency, and Berg-Kirkpatrick}]{cirik2022holm}
Volkan Cirik, Louis-Philippe Morency, and Taylor Berg-Kirkpatrick. 2022.
\newblock Holm: Hallucinating objects with language models for referring expression recognition in partially-observed scenes.
\newblock In \emph{Proceedings of the 60th Annual Meeting of the Association for Computational Linguistics (Volume 1: Long Papers)}, pages 5440--5453.

\bibitem[{Dahl et~al.(2024)Dahl, Magesh, Suzgun, and Ho}]{Dahl2024legal}
Matthew Dahl, Varun Magesh, Mirac Suzgun, and Daniel~E. Ho. 2024.
\newblock \href {https://hai.stanford.edu/news/hallucinating-law-legal-mistakes-large-language-models-are-pervasive?utm_source=linkedin&utm_medium=social&utm_content=Stanford+HAI_linkedin_HAI_202402221203_sf186435684&utm_campaign=&sf186435684=1} {Hallucinating law: Legal mistakes with large language models are pervasive}.

\bibitem[{Dai et~al.(2023)Dai, Liu, Ji, Su, and Fung}]{dai-etal-2023-plausible}
Wenliang Dai, Zihan Liu, Ziwei Ji, Dan Su, and Pascale Fung. 2023.
\newblock \href {https://doi.org/10.18653/v1/2023.eacl-main.156} {Plausible may not be faithful: Probing object hallucination in vision-language pre-training}.
\newblock In \emph{Proceedings of the 17th Conference of the European Chapter of the Association for Computational Linguistics}, pages 2136--2148, Dubrovnik, Croatia. Association for Computational Linguistics.

\bibitem[{Dale et~al.(2022)Dale, Voita, Barrault, and Costa-juss{\`a}}]{dale2022detecting}
David Dale, Elena Voita, Lo{\"\i}c Barrault, and Marta~R Costa-juss{\`a}. 2022.
\newblock Detecting and mitigating hallucinations in machine translation: Model internal workings alone do well, sentence similarity even better.
\newblock \emph{arXiv preprint arXiv:2212.08597}.

\bibitem[{Dale et~al.(2023)Dale, Voita, Lam, Hansanti, Ropers, Kalbassi, Gao, Barrault, and Costa-juss{\`a}}]{dale-etal-2023-halomi}
David Dale, Elena Voita, Janice Lam, Prangthip Hansanti, Christophe Ropers, Elahe Kalbassi, Cynthia Gao, Loic Barrault, and Marta Costa-juss{\`a}. 2023.
\newblock \href {https://doi.org/10.18653/v1/2023.emnlp-main.42} {{H}al{O}mi: A manually annotated benchmark for multilingual hallucination and omission detection in machine translation}.
\newblock In \emph{Proceedings of the 2023 Conference on Empirical Methods in Natural Language Processing}, pages 638--653, Singapore. Association for Computational Linguistics.

\bibitem[{Das et~al.(2022)Das, Saha, and Srihari}]{das-etal-2022-diving}
Souvik Das, Sougata Saha, and Rohini Srihari. 2022.
\newblock \href {https://doi.org/10.18653/v1/2022.findings-emnlp.48} {Diving deep into modes of fact hallucinations in dialogue systems}.
\newblock In \emph{Findings of the Association for Computational Linguistics: EMNLP 2022}, pages 684--699, Abu Dhabi, United Arab Emirates. Association for Computational Linguistics.

\bibitem[{Dong et~al.(2022)Dong, Wieting, and Verga}]{dong-etal-2022-faithful}
Yue Dong, John Wieting, and Pat Verga. 2022.
\newblock \href {https://doi.org/10.18653/v1/2022.findings-emnlp.76} {Faithful to the document or to the world? mitigating hallucinations via entity-linked knowledge in abstractive summarization}.
\newblock In \emph{Findings of the Association for Computational Linguistics: EMNLP 2022}, pages 1067--1082, Abu Dhabi, United Arab Emirates. Association for Computational Linguistics.

\bibitem[{Dziri et~al.(2022)Dziri, Kamalloo, Milton, Zaiane, Yu, Ponti, and Reddy}]{dziri2022faithdial}
Nouha Dziri, Ehsan Kamalloo, Sivan Milton, Osmar Zaiane, Mo~Yu, Edoardo~M Ponti, and Siva Reddy. 2022.
\newblock Faithdial: A faithful benchmark for information-seeking dialogue.
\newblock \emph{Transactions of the Association for Computational Linguistics}, 10:1473--1490.

\bibitem[{Fei et~al.(2023)Fei, Liu, Zhang, Zhang, and Chua}]{fei-etal-2023-scene}
Hao Fei, Qian Liu, Meishan Zhang, Min Zhang, and Tat-Seng Chua. 2023.
\newblock \href {https://doi.org/10.18653/v1/2023.acl-long.329} {Scene graph as pivoting: Inference-time image-free unsupervised multimodal machine translation with visual scene hallucination}.
\newblock In \emph{Proceedings of the 61st Annual Meeting of the Association for Computational Linguistics (Volume 1: Long Papers)}, pages 5980--5994, Toronto, Canada. Association for Computational Linguistics.

\bibitem[{Ferrando et~al.(2022)Ferrando, G{\'a}llego, Alastruey, Escolano, and Costa-juss{\`a}}]{ferrando-etal-2022-towards}
Javier Ferrando, Gerard~I. G{\'a}llego, Belen Alastruey, Carlos Escolano, and Marta~R. Costa-juss{\`a}. 2022.
\newblock \href {https://doi.org/10.18653/v1/2022.emnlp-main.599} {Towards opening the black box of neural machine translation: Source and target interpretations of the transformer}.
\newblock In \emph{Proceedings of the 2022 Conference on Empirical Methods in Natural Language Processing}, pages 8756--8769, Abu Dhabi, United Arab Emirates. Association for Computational Linguistics.

\bibitem[{Filippova(2020{\natexlab{a}})}]{filippova2020controlled}
Katja Filippova. 2020{\natexlab{a}}.
\newblock Controlled hallucinations: Learning to generate faithfully from noisy data.
\newblock \emph{arXiv preprint arXiv:2010.05873}.

\bibitem[{Filippova(2020{\natexlab{b}})}]{filippova-2020-controlled}
Katja Filippova. 2020{\natexlab{b}}.
\newblock \href {https://doi.org/10.18653/v1/2020.findings-emnlp.76} {Controlled hallucinations: Learning to generate faithfully from noisy data}.
\newblock In \emph{Findings of the Association for Computational Linguistics: EMNLP 2020}, pages 864--870, Online. Association for Computational Linguistics.

\bibitem[{Fink et~al.(2014)Fink, Benedek, Unterrainer, Papousek, and Weiss}]{fink2014creativity}
Andreas Fink, Mathias Benedek, Human-F Unterrainer, Ilona Papousek, and Elisabeth~M Weiss. 2014.
\newblock Creativity and psychopathology: are there similar mental processes involved in creativity and in psychosis-proneness?
\newblock \emph{Frontiers in psychology}, 5:117336.

\bibitem[{Flores and Cohan(2024)}]{flores-cohan-2024-benefits}
Lorenzo~Jaime Flores and Arman Cohan. 2024.
\newblock \href {https://aclanthology.org/2024.eacl-short.13} {On the benefits of fine-grained loss truncation: A case study on factuality in summarization}.
\newblock In \emph{Proceedings of the 18th Conference of the European Chapter of the Association for Computational Linguistics (Volume 2: Short Papers)}, pages 138--150, St. Julian{'}s, Malta. Association for Computational Linguistics.

\bibitem[{Friedl et~al.(2021)Friedl, Rizos, Stappen, Hasan, Specia, Hain, and Schuller}]{friedl-etal-2021-uncertainty}
Korbinian Friedl, Georgios Rizos, Lukas Stappen, Madina Hasan, Lucia Specia, Thomas Hain, and Bj{\"o}rn Schuller. 2021.
\newblock \href {https://doi.org/10.18653/v1/2021.findings-acl.443} {Uncertainty aware review hallucination for science article classification}.
\newblock In \emph{Findings of the Association for Computational Linguistics: ACL-IJCNLP 2021}, pages 5004--5009, Online. Association for Computational Linguistics.

\bibitem[{Gautam et~al.(2024)Gautam, Venkit, and Ghosh}]{gautam2024melting}
Sanjana Gautam, Pranav~Narayanan Venkit, and Sourojit Ghosh. 2024.
\newblock From melting pots to misrepresentations: Exploring harms in generative ai.
\newblock \emph{arXiv preprint arXiv:2403.10776}.

\bibitem[{Goldberg et~al.(2022)Goldberg, Kozareva, and Zhang}]{goldberg2022findings}
Yoav Goldberg, Zornitsa Kozareva, and Yue Zhang. 2022.
\newblock Findings of the association for computational linguistics: Emnlp 2022.
\newblock In \emph{Findings of the Association for Computational Linguistics: EMNLP 2022}.

\bibitem[{Gonz{\'a}lez~Corbelle et~al.(2022)Gonz{\'a}lez~Corbelle, Bugar{\'\i}n-Diz, Alonso-Moral, and Taboada}]{gonzalez-corbelle-etal-2022-dealing}
Javier Gonz{\'a}lez~Corbelle, Alberto Bugar{\'\i}n-Diz, Jose Alonso-Moral, and Juan Taboada. 2022.
\newblock \href {https://doi.org/10.18653/v1/2022.inlg-main.10} {Dealing with hallucination and omission in neural natural language generation: A use case on meteorology.}
\newblock In \emph{Proceedings of the 15th International Conference on Natural Language Generation}, pages 121--130, Waterville, Maine, USA and virtual meeting. Association for Computational Linguistics.

\bibitem[{Gonz{\'a}lez-Corbelle et~al.(2022)Gonz{\'a}lez-Corbelle, Diz, Alonso-Moral, and Taboada}]{gonzalez2022dealing}
Javier Gonz{\'a}lez-Corbelle, Alberto~Bugar{\'\i}n Diz, Jose Alonso-Moral, and Juan Taboada. 2022.
\newblock Dealing with hallucination and omission in neural natural language generation: A use case on meteorology.
\newblock In \emph{Proceedings of the 15th International Conference on Natural Language Generation}, pages 121--130.

\bibitem[{Guerreiro et~al.(2023{\natexlab{a}})Guerreiro, Colombo, Piantanida, and Martins}]{guerreiro-etal-2023-optimal}
Nuno~M. Guerreiro, Pierre Colombo, Pablo Piantanida, and Andr{\'e} Martins. 2023{\natexlab{a}}.
\newblock \href {https://doi.org/10.18653/v1/2023.acl-long.770} {Optimal transport for unsupervised hallucination detection in neural machine translation}.
\newblock In \emph{Proceedings of the 61st Annual Meeting of the Association for Computational Linguistics (Volume 1: Long Papers)}, pages 13766--13784, Toronto, Canada. Association for Computational Linguistics.

\bibitem[{Guerreiro et~al.(2023{\natexlab{b}})Guerreiro, Voita, and Martins}]{guerreiro-etal-2023-looking}
Nuno~M. Guerreiro, Elena Voita, and Andr{\'e} Martins. 2023{\natexlab{b}}.
\newblock \href {https://doi.org/10.18653/v1/2023.eacl-main.75} {Looking for a needle in a haystack: A comprehensive study of hallucinations in neural machine translation}.
\newblock In \emph{Proceedings of the 17th Conference of the European Chapter of the Association for Computational Linguistics}, pages 1059--1075, Dubrovnik, Croatia. Association for Computational Linguistics.

\bibitem[{Gupta et~al.(2024)Gupta, Venkit, Wilson, and Passonneau}]{gupta2024sociodemographic}
Vipul Gupta, Pranav~Narayanan Venkit, Shomir Wilson, and Rebecca~J. Passonneau. 2024.
\newblock \href {http://arxiv.org/abs/2306.08158} {Sociodemographic bias in language models: A survey and forward path}.

\bibitem[{Hsu et~al.(2010)Hsu, Lin, Hsu, Liao, and Yu}]{hsu_face_hallucination}
Chih-Chung Hsu, Chia-Wen Lin, Chiou-Ting Hsu, Hong-Yuan~Mark Liao, and Jen-Yu Yu. 2010.
\newblock \href {https://doi.org/10.1109/MMSP.2010.5662063} {Face hallucination using bayesian global estimation and local basis selection}.
\newblock In \emph{2010 IEEE International Workshop on Multimedia Signal Processing}, pages 449--453.

\bibitem[{Huang et~al.(2023)Huang, Yu, Ma, Zhong, Feng, Wang, Chen, Peng, Feng, Qin et~al.}]{huang2023survey}
Lei Huang, Weijiang Yu, Weitao Ma, Weihong Zhong, Zhangyin Feng, Haotian Wang, Qianglong Chen, Weihua Peng, Xiaocheng Feng, Bing Qin, et~al. 2023.
\newblock A survey on hallucination in large language models: Principles, taxonomy, challenges, and open questions.
\newblock \emph{arXiv preprint arXiv:2311.05232}.

\bibitem[{Irvine and Callison-Burch(2014)}]{irvine2014hallucinating}
Ann Irvine and Chris Callison-Burch. 2014.
\newblock Hallucinating phrase translations for low resource mt.
\newblock In \emph{Proceedings of the Eighteenth Conference on Computational Natural Language Learning}, pages 160--170.

\bibitem[{Obaid~ul Islam et~al.(2023)Obaid~ul Islam, {\v{S}}krjanec, Dusek, and Demberg}]{obaid-ul-islam-etal-2023-tackling}
Saad Obaid~ul Islam, Iza {\v{S}}krjanec, Ondrej Dusek, and Vera Demberg. 2023.
\newblock \href {https://doi.org/10.18653/v1/2023.inlg-main.30} {Tackling hallucinations in neural chart summarization}.
\newblock In \emph{Proceedings of the 16th International Natural Language Generation Conference}, pages 414--423, Prague, Czechia. Association for Computational Linguistics.

\bibitem[{Islam et~al.(2023)Islam, {\v{S}}krjanec, Du{\v{s}}ek, and Demberg}]{islam2023tackling}
Saad Obaid~Ul Islam, Iza {\v{S}}krjanec, Ond{\v{r}}ej Du{\v{s}}ek, and Vera Demberg. 2023.
\newblock Tackling hallucinations in neural chart summarization.
\newblock In \emph{Proceedings of the 16th International Natural Language Generation Conference}, pages 414--423.

\bibitem[{Ji et~al.(2023{\natexlab{a}})Ji, Lee, Frieske, Yu, Su, Xu, Ishii, Bang, Madotto, and Fung}]{ji2023survey}
Ziwei Ji, Nayeon Lee, Rita Frieske, Tiezheng Yu, Dan Su, Yan Xu, Etsuko Ishii, Ye~Jin Bang, Andrea Madotto, and Pascale Fung. 2023{\natexlab{a}}.
\newblock Survey of hallucination in natural language generation.
\newblock \emph{ACM Computing Surveys}, 55(12):1--38.

\bibitem[{Ji et~al.(2023{\natexlab{b}})Ji, Liu, Lee, Yu, Wilie, Zeng, and Fung}]{ji-etal-2023-rho}
Ziwei Ji, Zihan Liu, Nayeon Lee, Tiezheng Yu, Bryan Wilie, Min Zeng, and Pascale Fung. 2023{\natexlab{b}}.
\newblock \href {https://doi.org/10.18653/v1/2023.findings-acl.275} {{RHO}: Reducing hallucination in open-domain dialogues with knowledge grounding}.
\newblock In \emph{Findings of the Association for Computational Linguistics: ACL 2023}, pages 4504--4522, Toronto, Canada. Association for Computational Linguistics.

\bibitem[{Ji et~al.(2023{\natexlab{c}})Ji, Tiezheng, Xu, Lee, Ishii, and Fung}]{ji2023towards}
Ziwei Ji, YU~Tiezheng, Yan Xu, Nayeon Lee, Etsuko Ishii, and Pascale Fung. 2023{\natexlab{c}}.
\newblock Towards mitigating llm hallucination via self reflection.
\newblock In \emph{The 2023 Conference on Empirical Methods in Natural Language Processing}.

\bibitem[{Ji et~al.(2023{\natexlab{d}})Ji, Yu, Xu, Lee, Ishii, and Fung}]{ji-etal-2023-towards}
Ziwei Ji, Tiezheng Yu, Yan Xu, Nayeon Lee, Etsuko Ishii, and Pascale Fung. 2023{\natexlab{d}}.
\newblock \href {https://doi.org/10.18653/v1/2023.findings-emnlp.123} {Towards mitigating {LLM} hallucination via self reflection}.
\newblock In \emph{Findings of the Association for Computational Linguistics: EMNLP 2023}, pages 1827--1843, Singapore. Association for Computational Linguistics.

\bibitem[{Jian et~al.(2022)Jian, Gao, and Vosoughi}]{jian-etal-2022-embedding}
Yiren Jian, Chongyang Gao, and Soroush Vosoughi. 2022.
\newblock \href {https://doi.org/10.18653/v1/2022.naacl-main.404} {Embedding hallucination for few-shot language fine-tuning}.
\newblock In \emph{Proceedings of the 2022 Conference of the North American Chapter of the Association for Computational Linguistics: Human Language Technologies}, pages 5522--5530, Seattle, United States. Association for Computational Linguistics.

\bibitem[{Jin and Mihalcea(2022)}]{jin2022natural}
Zhijing Jin and Rada Mihalcea. 2022.
\newblock Natural language processing for policymaking.
\newblock In \emph{Handbook of Computational Social Science for Policy}, pages 141--162. Springer International Publishing Cham.

\bibitem[{Karpathy(2015)}]{karpathy2015unreasonable}
Andrej Karpathy. 2015.
\newblock The unreasonable effectiveness of recurrent neural networks.
\newblock \url{https://karpathy.github.io/2015/05/21/rnn-effectiveness/}.

\bibitem[{Kothyari et~al.(2023)Kothyari, Dhingra, Sarawagi, and Chakrabarti}]{kothyari-etal-2023-crush4sql}
Mayank Kothyari, Dhruva Dhingra, Sunita Sarawagi, and Soumen Chakrabarti. 2023.
\newblock \href {https://doi.org/10.18653/v1/2023.emnlp-main.868} {{CRUSH}4{SQL}: Collective retrieval using schema hallucination for {T}ext2{SQL}}.
\newblock In \emph{Proceedings of the 2023 Conference on Empirical Methods in Natural Language Processing}, pages 14054--14066, Singapore. Association for Computational Linguistics.

\bibitem[{Ladhak et~al.(2022)Ladhak, Durmus, and Hashimoto}]{ladhak2022contrastive}
Faisal Ladhak, Esin Durmus, and Tatsunori Hashimoto. 2022.
\newblock Contrastive error attribution for finetuned language models.
\newblock \emph{arXiv preprint arXiv:2212.10722}.

\bibitem[{Ladhak et~al.(2023)Ladhak, Durmus, Suzgun, Zhang, Jurafsky, McKeown, and Hashimoto}]{ladhak2023pre}
Faisal Ladhak, Esin Durmus, Mirac Suzgun, Tianyi Zhang, Dan Jurafsky, Kathleen McKeown, and Tatsunori~B Hashimoto. 2023.
\newblock When do pre-training biases propagate to downstream tasks? a case study in text summarization.
\newblock In \emph{Proceedings of the 17th Conference of the European Chapter of the Association for Computational Linguistics}, pages 3206--3219.

\bibitem[{Lango and Dusek(2023)}]{lango-dusek-2023-critic}
Mateusz Lango and Ondrej Dusek. 2023.
\newblock \href {https://doi.org/10.18653/v1/2023.emnlp-main.172} {Critic-driven decoding for mitigating hallucinations in data-to-text generation}.
\newblock In \emph{Proceedings of the 2023 Conference on Empirical Methods in Natural Language Processing}, pages 2853--2862, Singapore. Association for Computational Linguistics.

\bibitem[{Legault(2020)}]{legault2020encyclopedia}
Lisa Legault. 2020.
\newblock Encyclopedia of personality and individual differences.
\newblock \emph{Encyclopedia of Personality and Individual Differences}, pages 1--5.

\bibitem[{Li et~al.(2023)Li, Du, Zhou, Wang, Zhao, and Wen}]{li-etal-2023-evaluating}
Yifan Li, Yifan Du, Kun Zhou, Jinpeng Wang, Xin Zhao, and Ji-Rong Wen. 2023.
\newblock \href {https://doi.org/10.18653/v1/2023.emnlp-main.20} {Evaluating object hallucination in large vision-language models}.
\newblock In \emph{Proceedings of the 2023 Conference on Empirical Methods in Natural Language Processing}, pages 292--305, Singapore. Association for Computational Linguistics.

\bibitem[{Liu and Wan(2023)}]{liu-wan-2023-models}
Hui Liu and Xiaojun Wan. 2023.
\newblock \href {https://doi.org/10.18653/v1/2023.emnlp-main.723} {Models see hallucinations: Evaluating the factuality in video captioning}.
\newblock In \emph{Proceedings of the 2023 Conference on Empirical Methods in Natural Language Processing}, pages 11807--11823, Singapore. Association for Computational Linguistics.

\bibitem[{Liu et~al.(2022)Liu, Zhang, Brockett, Mao, Sui, Chen, and Dolan}]{liu-etal-2022-token}
Tianyu Liu, Yizhe Zhang, Chris Brockett, Yi~Mao, Zhifang Sui, Weizhu Chen, and Bill Dolan. 2022.
\newblock \href {https://doi.org/10.18653/v1/2022.acl-long.464} {A token-level reference-free hallucination detection benchmark for free-form text generation}.
\newblock In \emph{Proceedings of the 60th Annual Meeting of the Association for Computational Linguistics (Volume 1: Long Papers)}, pages 6723--6737, Dublin, Ireland. Association for Computational Linguistics.

\bibitem[{Longpre et~al.(2021)Longpre, Perisetla, Chen, Ramesh, DuBois, and Singh}]{longpre-etal-2021-entity}
Shayne Longpre, Kartik Perisetla, Anthony Chen, Nikhil Ramesh, Chris DuBois, and Sameer Singh. 2021.
\newblock \href {https://doi.org/10.18653/v1/2021.emnlp-main.565} {Entity-based knowledge conflicts in question answering}.
\newblock In \emph{Proceedings of the 2021 Conference on Empirical Methods in Natural Language Processing}, pages 7052--7063, Online and Punta Cana, Dominican Republic. Association for Computational Linguistics.

\bibitem[{Maharaj et~al.(2023)Maharaj, Saxena, Kumar, Mishra, and Bhattacharyya}]{maharaj-etal-2023-eyes}
Kishan Maharaj, Ashita Saxena, Raja Kumar, Abhijit Mishra, and Pushpak Bhattacharyya. 2023.
\newblock \href {https://doi.org/10.18653/v1/2023.findings-emnlp.764} {Eyes show the way: Modelling gaze behaviour for hallucination detection}.
\newblock In \emph{Findings of the Association for Computational Linguistics: EMNLP 2023}, pages 11424--11438, Singapore. Association for Computational Linguistics.

\bibitem[{Maheshwari et~al.(2023)Maheshwari, Shekhar, Saxena, and Chhaya}]{maheshwari-etal-2023-open}
Himanshu Maheshwari, Sumit Shekhar, Apoorv Saxena, and Niyati Chhaya. 2023.
\newblock \href {https://doi.org/10.18653/v1/2023.findings-acl.151} {Open-world factually consistent question generation}.
\newblock In \emph{Findings of the Association for Computational Linguistics: ACL 2023}, pages 2390--2404, Toronto, Canada. Association for Computational Linguistics.

\bibitem[{Maleki et~al.(2024)Maleki, Padmanabhan, and Dutta}]{maleki2024ai}
Negar Maleki, Balaji Padmanabhan, and Kaushik Dutta. 2024.
\newblock Ai hallucinations: A misnomer worth clarifying.
\newblock \emph{arXiv preprint arXiv:2401.06796}.

\bibitem[{Manakul et~al.(2023)Manakul, Liusie, and Gales}]{manakul-etal-2023-selfcheckgpt}
Potsawee Manakul, Adian Liusie, and Mark Gales. 2023.
\newblock \href {https://doi.org/10.18653/v1/2023.emnlp-main.557} {{S}elf{C}heck{GPT}: Zero-resource black-box hallucination detection for generative large language models}.
\newblock In \emph{Proceedings of the 2023 Conference on Empirical Methods in Natural Language Processing}, pages 9004--9017, Singapore. Association for Computational Linguistics.

\bibitem[{Marfurt and Henderson(2022)}]{marfurt-henderson-2022-unsupervised}
Andreas Marfurt and James Henderson. 2022.
\newblock \href {https://doi.org/10.18653/v1/2022.gem-1.21} {Unsupervised token-level hallucination detection from summary generation by-products}.
\newblock In \emph{Proceedings of the 2nd Workshop on Natural Language Generation, Evaluation, and Metrics (GEM)}, pages 248--261, Abu Dhabi, United Arab Emirates (Hybrid). Association for Computational Linguistics.

\bibitem[{Mason et~al.(2021)Mason, Kuypers, Reckweg, M{\"u}ller, Tse, Da~Rios, Toennes, Stiers, Feilding, and Ramaekers}]{mason2021spontaneous}
NL~Mason, KPC Kuypers, JT~Reckweg, F~M{\"u}ller, DHY Tse, B~Da~Rios, SW~Toennes, P~Stiers, A~Feilding, and JG~Ramaekers. 2021.
\newblock Spontaneous and deliberate creative cognition during and after psilocybin exposure.
\newblock \emph{Translational psychiatry}, 11(1):209.

\bibitem[{Massarelli et~al.(2020)Massarelli, Petroni, Piktus, Ott, Rockt{\"a}schel, Plachouras, Silvestri, and Riedel}]{massarelli-etal-2020-decoding}
Luca Massarelli, Fabio Petroni, Aleksandra Piktus, Myle Ott, Tim Rockt{\"a}schel, Vassilis Plachouras, Fabrizio Silvestri, and Sebastian Riedel. 2020.
\newblock \href {https://doi.org/10.18653/v1/2020.findings-emnlp.22} {How decoding strategies affect the verifiability of generated text}.
\newblock In \emph{Findings of the Association for Computational Linguistics: EMNLP 2020}, pages 223--235, Online. Association for Computational Linguistics.

\bibitem[{Maynez et~al.(2020)Maynez, Narayan, Bohnet, and McDonald}]{maynez-etal-2020-faithfulness}
Joshua Maynez, Shashi Narayan, Bernd Bohnet, and Ryan McDonald. 2020.
\newblock \href {https://doi.org/10.18653/v1/2020.acl-main.173} {On faithfulness and factuality in abstractive summarization}.
\newblock In \emph{Proceedings of the 58th Annual Meeting of the Association for Computational Linguistics}, pages 1906--1919, Online. Association for Computational Linguistics.

\bibitem[{McGowan et~al.(2023)McGowan, Gui, Dobbs, Shuster, Cotter, Selloni, Goodman, Srivastava, Cecchi, and Corcoran}]{mcgowan2023chatgpt}
Alessia McGowan, Yunlai Gui, Matthew Dobbs, Sophia Shuster, Matthew Cotter, Alexandria Selloni, Marianne Goodman, Agrima Srivastava, Guillermo~A Cecchi, and Cheryl~M Corcoran. 2023.
\newblock Chatgpt and bard exhibit spontaneous citation fabrication during psychiatry literature search.
\newblock \emph{Psychiatry Research}, 326:115334.

\bibitem[{McKenna et~al.(2023)McKenna, Li, Cheng, Hosseini, Johnson, and Steedman}]{mckenna-etal-2023-sources}
Nick McKenna, Tianyi Li, Liang Cheng, Mohammad Hosseini, Mark Johnson, and Mark Steedman. 2023.
\newblock \href {https://doi.org/10.18653/v1/2023.findings-emnlp.182} {Sources of hallucination by large language models on inference tasks}.
\newblock In \emph{Findings of the Association for Computational Linguistics: EMNLP 2023}, pages 2758--2774, Singapore. Association for Computational Linguistics.

\bibitem[{Mielke et~al.(2022)Mielke, Szlam, Dinan, and Boureau}]{mielke2022reducing}
Sabrina~J Mielke, Arthur Szlam, Emily Dinan, and Y-Lan Boureau. 2022.
\newblock Reducing conversational agents’ overconfidence through linguistic calibration.
\newblock \emph{Transactions of the Association for Computational Linguistics}, 10:857--872.

\bibitem[{Millidge(2023)}]{millidge2023llms}
Beren Millidge. 2023.
\newblock \href {https://www.beren.io/2023-03-19-LLMs-confabulate-not-hallucinate/} {Llms confabulate not hallucinate}.

\bibitem[{M{\"u}ller et~al.(2020)M{\"u}ller, Rios, and Sennrich}]{muller-etal-2020-domain}
Mathias M{\"u}ller, Annette Rios, and Rico Sennrich. 2020.
\newblock \href {https://aclanthology.org/2020.amta-research.14} {Domain robustness in neural machine translation}.
\newblock In \emph{Proceedings of the 14th Conference of the Association for Machine Translation in the Americas (Volume 1: Research Track)}, pages 151--164, Virtual. Association for Machine Translation in the Americas.

\bibitem[{Nan et~al.(2021)Nan, Nallapati, Wang, Nogueira~dos Santos, Zhu, Zhang, McKeown, and Xiang}]{nan-etal-2021-entity}
Feng Nan, Ramesh Nallapati, Zhiguo Wang, Cicero Nogueira~dos Santos, Henghui Zhu, Dejiao Zhang, Kathleen McKeown, and Bing Xiang. 2021.
\newblock \href {https://doi.org/10.18653/v1/2021.eacl-main.235} {Entity-level factual consistency of abstractive text summarization}.
\newblock In \emph{Proceedings of the 16th Conference of the European Chapter of the Association for Computational Linguistics: Main Volume}, pages 2727--2733, Online. Association for Computational Linguistics.

\bibitem[{Narayanan~Venkit(2023)}]{narayanan2023towards}
Pranav Narayanan~Venkit. 2023.
\newblock Towards a holistic approach: Understanding sociodemographic biases in nlp models using an interdisciplinary lens.
\newblock In \emph{Proceedings of the 2023 AAAI/ACM Conference on AI, Ethics, and Society}, pages 1004--1005.

\bibitem[{Narayanan~Venkit et~al.(2023)Narayanan~Venkit, Gautam, Panchanadikar, Huang, and Wilson}]{narayanan2023unmasking}
Pranav Narayanan~Venkit, Sanjana Gautam, Ruchi Panchanadikar, Ting-Hao Huang, and Shomir Wilson. 2023.
\newblock Unmasking nationality bias: A study of human perception of nationalities in ai-generated articles.
\newblock In \emph{Proceedings of the 2023 AAAI/ACM Conference on AI, Ethics, and Society}, pages 554--565.

\bibitem[{News(2023)}]{us_ranking}
US~News. 2023.
\newblock \href {https://www.usnews.com/best-colleges/rankings/national-universities} {Best national university rankings 2023}.

\bibitem[{Nie et~al.(2019)Nie, Yao, Wang, Pan, and Lin}]{nie-etal-2019-simple}
Feng Nie, Jin-Ge Yao, Jinpeng Wang, Rong Pan, and Chin-Yew Lin. 2019.
\newblock \href {https://doi.org/10.18653/v1/P19-1256} {A simple recipe towards reducing hallucination in neural surface realisation}.
\newblock In \emph{Proceedings of the 57th Annual Meeting of the Association for Computational Linguistics}, pages 2673--2679, Florence, Italy. Association for Computational Linguistics.

\bibitem[{Pal et~al.(2023)Pal, Umapathi, and Sankarasubbu}]{pal-etal-2023-med}
Ankit Pal, Logesh~Kumar Umapathi, and Malaikannan Sankarasubbu. 2023.
\newblock \href {https://doi.org/10.18653/v1/2023.conll-1.21} {{M}ed-{HALT}: Medical domain hallucination test for large language models}.
\newblock In \emph{Proceedings of the 27th Conference on Computational Natural Language Learning (CoNLL)}, pages 314--334, Singapore. Association for Computational Linguistics.

\bibitem[{Pfeiffer et~al.(2023)Pfeiffer, Piccinno, Nicosia, Wang, Reid, and Ruder}]{pfeiffer-etal-2023-mmt5}
Jonas Pfeiffer, Francesco Piccinno, Massimo Nicosia, Xinyi Wang, Machel Reid, and Sebastian Ruder. 2023.
\newblock \href {https://doi.org/10.18653/v1/2023.findings-emnlp.132} {mm{T}5: Modular multilingual pre-training solves source language hallucinations}.
\newblock In \emph{Findings of the Association for Computational Linguistics: EMNLP 2023}, pages 1978--2008, Singapore. Association for Computational Linguistics.

\bibitem[{van~der Poel et~al.(2022)van~der Poel, Cotterell, and Meister}]{van-der-poel-etal-2022-mutual}
Liam van~der Poel, Ryan Cotterell, and Clara Meister. 2022.
\newblock \href {https://doi.org/10.18653/v1/2022.emnlp-main.399} {Mutual information alleviates hallucinations in abstractive summarization}.
\newblock In \emph{Proceedings of the 2022 Conference on Empirical Methods in Natural Language Processing}, pages 5956--5965, Abu Dhabi, United Arab Emirates. Association for Computational Linguistics.

\bibitem[{Polat et~al.(2023)Polat, Tiddi, Groth, and Vossen}]{polat2023improving}
Fina Polat, Ilaria Tiddi, Paul Groth, and Piek Vossen. 2023.
\newblock Improving graph-to-text generation using cycle training.
\newblock In \emph{Proceedings of the 4th Conference on Language, Data and Knowledge}, pages 256--261.

\bibitem[{Qiu et~al.(2023)Qiu, Ziser, Korhonen, Ponti, and Cohen}]{qiu-etal-2023-detecting}
Yifu Qiu, Yftah Ziser, Anna Korhonen, Edoardo Ponti, and Shay Cohen. 2023.
\newblock \href {https://doi.org/10.18653/v1/2023.emnlp-main.551} {Detecting and mitigating hallucinations in multilingual summarisation}.
\newblock In \emph{Proceedings of the 2023 Conference on Empirical Methods in Natural Language Processing}, pages 8914--8932, Singapore. Association for Computational Linguistics.

\bibitem[{Ramakrishna et~al.(2023)Ramakrishna, Gupta, Lehmann, and Ziyadi}]{ramakrishna2023invite}
Anil Ramakrishna, Rahul Gupta, Jens Lehmann, and Morteza Ziyadi. 2023.
\newblock Invite: a testbed of automatically generated invalid questions to evaluate large language models for hallucinations.
\newblock In \emph{The 2023 Conference on Empirical Methods in Natural Language Processing}.

\bibitem[{Raunak et~al.(2021)Raunak, Menezes, and Junczys-Dowmunt}]{raunak-etal-2021-curious}
Vikas Raunak, Arul Menezes, and Marcin Junczys-Dowmunt. 2021.
\newblock \href {https://doi.org/10.18653/v1/2021.naacl-main.92} {The curious case of hallucinations in neural machine translation}.
\newblock In \emph{Proceedings of the 2021 Conference of the North American Chapter of the Association for Computational Linguistics: Human Language Technologies}, pages 1172--1183, Online. Association for Computational Linguistics.

\bibitem[{Rawte et~al.(2023{\natexlab{a}})Rawte, Chakraborty, Pathak, Sarkar, Tonmoy, Chadha, Sheth, and Das}]{rawte2023troubling}
Vipula Rawte, Swagata Chakraborty, Agnibh Pathak, Anubhav Sarkar, SM~Tonmoy, Aman Chadha, Amit~P Sheth, and Amitava Das. 2023{\natexlab{a}}.
\newblock The troubling emergence of hallucination in large language models--an extensive definition, quantification, and prescriptive remediations.
\newblock \emph{arXiv preprint arXiv:2310.04988}.

\bibitem[{Rawte et~al.(2023{\natexlab{b}})Rawte, Sheth, and Das}]{rawte2023survey}
Vipula Rawte, Amit Sheth, and Amitava Das. 2023{\natexlab{b}}.
\newblock A survey of hallucination in large foundation models.
\newblock \emph{arXiv preprint arXiv:2309.05922}.

\bibitem[{Rohrbach et~al.(2018)Rohrbach, Hendricks, Burns, Darrell, and Saenko}]{rohrbach-etal-2018-object}
Anna Rohrbach, Lisa~Anne Hendricks, Kaylee Burns, Trevor Darrell, and Kate Saenko. 2018.
\newblock \href {https://doi.org/10.18653/v1/D18-1437} {Object hallucination in image captioning}.
\newblock In \emph{Proceedings of the 2018 Conference on Empirical Methods in Natural Language Processing}, pages 4035--4045, Brussels, Belgium. Association for Computational Linguistics.

\bibitem[{Roller et~al.(2021)Roller, Dinan, Goyal, Ju, Williamson, Liu, Xu, Ott, Smith, Boureau, and Weston}]{roller-etal-2021-recipes}
Stephen Roller, Emily Dinan, Naman Goyal, Da~Ju, Mary Williamson, Yinhan Liu, Jing Xu, Myle Ott, Eric~Michael Smith, Y-Lan Boureau, and Jason Weston. 2021.
\newblock \href {https://doi.org/10.18653/v1/2021.eacl-main.24} {Recipes for building an open-domain chatbot}.
\newblock In \emph{Proceedings of the 16th Conference of the European Chapter of the Association for Computational Linguistics: Main Volume}, pages 300--325, Online. Association for Computational Linguistics.

\bibitem[{Rosen et~al.(2013)Rosen, Whaling, Carrier, Cheever, and Rokkum}]{rosen2013media}
Larry~D Rosen, Kelly Whaling, L~Mark Carrier, Nancy~A Cheever, and Jeffrey Rokkum. 2013.
\newblock The media and technology usage and attitudes scale: An empirical investigation.
\newblock \emph{Computers in human behavior}, 29(6):2501--2511.

\bibitem[{Sadat et~al.(2023)Sadat, Zhou, Lange, Araki, Gundroo, Wang, Menon, Parvez, and Feng}]{sadat-etal-2023-delucionqa}
Mobashir Sadat, Zhengyu Zhou, Lukas Lange, Jun Araki, Arsalan Gundroo, Bingqing Wang, Rakesh Menon, Md~Parvez, and Zhe Feng. 2023.
\newblock \href {https://doi.org/10.18653/v1/2023.findings-emnlp.59} {{D}elucion{QA}: Detecting hallucinations in domain-specific question answering}.
\newblock In \emph{Findings of the Association for Computational Linguistics: EMNLP 2023}, pages 822--835, Singapore. Association for Computational Linguistics.

\bibitem[{Samir and Silfverberg(2022)}]{samir-silfverberg-2022-one}
Farhan Samir and Miikka Silfverberg. 2022.
\newblock \href {https://doi.org/10.18653/v1/2022.computel-1.5} {One wug, two wug+s transformer inflection models hallucinate affixes}.
\newblock In \emph{Proceedings of the Fifth Workshop on the Use of Computational Methods in the Study of Endangered Languages}, pages 31--40, Dublin, Ireland. Association for Computational Linguistics.

\bibitem[{Shen et~al.(2023)Shen, Xuan, and Liang}]{shen-etal-2023-mitigating}
Jianbin Shen, Junyu Xuan, and Christy Liang. 2023.
\newblock \href {https://doi.org/10.18653/v1/2023.findings-emnlp.1059} {Mitigating intrinsic named entity-related hallucinations of abstractive text summarization}.
\newblock In \emph{Findings of the Association for Computational Linguistics: EMNLP 2023}, pages 15807--15824, Singapore. Association for Computational Linguistics.

\bibitem[{Shi et~al.(2023)Shi, Zhu, Zhang, and Li}]{shi-etal-2023-hallucination}
Xiao Shi, Zhengyuan Zhu, Zeyu Zhang, and Chengkai Li. 2023.
\newblock \href {https://doi.org/10.18653/v1/2023.emnlp-main.770} {Hallucination mitigation in natural language generation from large-scale open-domain knowledge graphs}.
\newblock In \emph{Proceedings of the 2023 Conference on Empirical Methods in Natural Language Processing}, pages 12506--12521, Singapore. Association for Computational Linguistics.

\bibitem[{Shuster et~al.(2021)Shuster, Poff, Chen, Kiela, and Weston}]{shuster-etal-2021-retrieval-augmentation}
Kurt Shuster, Spencer Poff, Moya Chen, Douwe Kiela, and Jason Weston. 2021.
\newblock \href {https://doi.org/10.18653/v1/2021.findings-emnlp.320} {Retrieval augmentation reduces hallucination in conversation}.
\newblock In \emph{Findings of the Association for Computational Linguistics: EMNLP 2021}, pages 3784--3803, Punta Cana, Dominican Republic. Association for Computational Linguistics.

\bibitem[{Slobodkin et~al.(2023)Slobodkin, Goldman, Caciularu, Dagan, and Ravfogel}]{slobodkin2023curious}
Aviv Slobodkin, Omer Goldman, Avi Caciularu, Ido Dagan, and Shauli Ravfogel. 2023.
\newblock The curious case of hallucinatory (un) answerability: Finding truths in the hidden states of over-confident large language models.
\newblock In \emph{Proceedings of the 2023 Conference on Empirical Methods in Natural Language Processing}, pages 3607--3625.

\bibitem[{Smith et~al.(2022)Smith, Hsu, Qian, Roller, Boureau, and Weston}]{smith-etal-2022-human}
Eric Smith, Orion Hsu, Rebecca Qian, Stephen Roller, Y-Lan Boureau, and Jason Weston. 2022.
\newblock \href {https://doi.org/10.18653/v1/2022.nlp4convai-1.8} {Human evaluation of conversations is an open problem: comparing the sensitivity of various methods for evaluating dialogue agents}.
\newblock In \emph{Proceedings of the 4th Workshop on NLP for Conversational AI}, pages 77--97, Dublin, Ireland. Association for Computational Linguistics.

\bibitem[{Son et~al.(2022)Son, Park, Hwang, Lee, Noh, and Lee}]{son-etal-2022-harim}
Seonil~(Simon) Son, Junsoo Park, Jeong-in Hwang, Junghwa Lee, Hyungjong Noh, and Yeonsoo Lee. 2022.
\newblock \href {https://aclanthology.org/2022.aacl-main.66} {{H}a{R}i{M}$^+$: Evaluating summary quality with hallucination risk}.
\newblock In \emph{Proceedings of the 2nd Conference of the Asia-Pacific Chapter of the Association for Computational Linguistics and the 12th International Joint Conference on Natural Language Processing (Volume 1: Long Papers)}, pages 895--924, Online only. Association for Computational Linguistics.

\bibitem[{Steele(2017)}]{steele2017hallucination}
Brent~J Steele. 2017.
\newblock Hallucination and intervention.
\newblock \emph{Global Discourse}, 7(2-3):201--218.

\bibitem[{Sun et~al.(2023)Sun, Li, Mi, Bie, Li, and Li}]{sun-etal-2023-towards}
Bin Sun, Yitong Li, Fei Mi, Fanhu Bie, Yiwei Li, and Kan Li. 2023.
\newblock \href {https://doi.org/10.18653/v1/2023.acl-short.148} {Towards fewer hallucinations in knowledge-grounded dialogue generation via augmentative and contrastive knowledge-dialogue}.
\newblock In \emph{Proceedings of the 61st Annual Meeting of the Association for Computational Linguistics (Volume 2: Short Papers)}, pages 1741--1750, Toronto, Canada. Association for Computational Linguistics.

\bibitem[{Sundar and Heck(2023)}]{sundar-heck-2023-ctbls}
Anirudh~S. Sundar and Larry Heck. 2023.
\newblock \href {https://doi.org/10.18653/v1/2023.nlp4convai-1.6} {c{TBLS}: Augmenting large language models with conversational tables}.
\newblock In \emph{Proceedings of the 5th Workshop on NLP for Conversational AI (NLP4ConvAI 2023)}, pages 59--70, Toronto, Canada. Association for Computational Linguistics.

\bibitem[{Terry et~al.(2017)Terry, Hayfield, Clarke, and Braun}]{terry2017thematic}
Gareth Terry, Nikki Hayfield, Victoria Clarke, and Virginia Braun. 2017.
\newblock Thematic analysis.
\newblock \emph{The SAGE handbook of qualitative research in psychology}, 2:17--37.

\bibitem[{Testoni and Bernardi(2021)}]{testoni-bernardi-2021-ive}
Alberto Testoni and Raffaella Bernardi. 2021.
\newblock \href {https://doi.org/10.18653/v1/2021.acl-srw.11} {{``}{I}{'}ve seen things you people wouldn{'}t believe{''}: Hallucinating entities in {G}uess{W}hat?!}
\newblock In \emph{Proceedings of the 59th Annual Meeting of the Association for Computational Linguistics and the 11th International Joint Conference on Natural Language Processing: Student Research Workshop}, pages 101--111, Online. Association for Computational Linguistics.

\bibitem[{Tonmoy et~al.(2024)Tonmoy, Zaman, Jain, Rani, Rawte, Chadha, and Das}]{tonmoy2024comprehensive}
SM~Tonmoy, SM~Zaman, Vinija Jain, Anku Rani, Vipula Rawte, Aman Chadha, and Amitava Das. 2024.
\newblock A comprehensive survey of hallucination mitigation techniques in large language models.
\newblock \emph{arXiv preprint arXiv:2401.01313}.

\bibitem[{Vaismoradi et~al.(2013)Vaismoradi, Turunen, and Bondas}]{vaismoradi2013content}
Mojtaba Vaismoradi, Hannele Turunen, and Terese Bondas. 2013.
\newblock Content analysis and thematic analysis: Implications for conducting a qualitative descriptive study.
\newblock \emph{Nursing \& health sciences}, 15(3):398--405.

\bibitem[{Van~Noorden and Perkel(2023)}]{van2023ai}
Richard Van~Noorden and Jeffrey~M Perkel. 2023.
\newblock Ai and science: what 1,600 researchers think.
\newblock \emph{Nature}, 621(7980):672--675.

\bibitem[{Venkit et~al.(2023)Venkit, Srinath, Gautam, Venkatraman, Gupta, Passonneau, and Wilson}]{venkit2023sentiment}
Pranav Venkit, Mukund Srinath, Sanjana Gautam, Saranya Venkatraman, Vipul Gupta, Rebecca~J Passonneau, and Shomir Wilson. 2023.
\newblock The sentiment problem: A critical survey towards deconstructing sentiment analysis.
\newblock In \emph{Proceedings of the 2023 Conference on Empirical Methods in Natural Language Processing}, pages 13743--13763.

\bibitem[{Vu et~al.(2022)Vu, Barua, Lester, Cer, Iyyer, and Constant}]{vu-etal-2022-overcoming}
Tu~Vu, Aditya Barua, Brian Lester, Daniel Cer, Mohit Iyyer, and Noah Constant. 2022.
\newblock \href {https://doi.org/10.18653/v1/2022.emnlp-main.630} {Overcoming catastrophic forgetting in zero-shot cross-lingual generation}.
\newblock In \emph{Proceedings of the 2022 Conference on Empirical Methods in Natural Language Processing}, pages 9279--9300, Abu Dhabi, United Arab Emirates. Association for Computational Linguistics.

\bibitem[{Waldendorf et~al.(2024)Waldendorf, Haddow, and Birch}]{waldendorf-etal-2024-contrastive}
Jonas Waldendorf, Barry Haddow, and Alexandra Birch. 2024.
\newblock \href {https://aclanthology.org/2024.eacl-long.155} {Contrastive decoding reduces hallucinations in large multilingual machine translation models}.
\newblock In \emph{Proceedings of the 18th Conference of the European Chapter of the Association for Computational Linguistics (Volume 1: Long Papers)}, pages 2526--2539, St. Julian{'}s, Malta. Association for Computational Linguistics.

\bibitem[{Wang and Sennrich(2020)}]{wang-sennrich-2020-exposure}
Chaojun Wang and Rico Sennrich. 2020.
\newblock \href {https://doi.org/10.18653/v1/2020.acl-main.326} {On exposure bias, hallucination and domain shift in neural machine translation}.
\newblock In \emph{Proceedings of the 58th Annual Meeting of the Association for Computational Linguistics}, pages 3544--3552, Online. Association for Computational Linguistics.

\bibitem[{Weller et~al.(2024)Weller, Marone, Weir, Lawrie, Khashabi, and Van~Durme}]{weller-etal-2024-according}
Orion Weller, Marc Marone, Nathaniel Weir, Dawn Lawrie, Daniel Khashabi, and Benjamin Van~Durme. 2024.
\newblock \href {https://aclanthology.org/2024.eacl-long.140} {{``}according to . . . {''}: Prompting language models improves quoting from pre-training data}.
\newblock In \emph{Proceedings of the 18th Conference of the European Chapter of the Association for Computational Linguistics (Volume 1: Long Papers)}, pages 2288--2301, St. Julian{'}s, Malta. Association for Computational Linguistics.

\bibitem[{Werning(2024)}]{werning2024generative}
Stefan Werning. 2024.
\newblock Generative ai and the technological imaginary of game design.
\newblock In \emph{Creative Tools and the Softwarization of Cultural Production}, pages 67--90. Springer.

\bibitem[{Wu et~al.(2024)Wu, Waheed, Zhang, Abdul-Mageed, and Aji}]{wu-etal-2024-lamini}
Minghao Wu, Abdul Waheed, Chiyu Zhang, Muhammad Abdul-Mageed, and Alham Aji. 2024.
\newblock \href {https://aclanthology.org/2024.eacl-long.57} {{L}a{M}ini-{LM}: A diverse herd of distilled models from large-scale instructions}.
\newblock In \emph{Proceedings of the 18th Conference of the European Chapter of the Association for Computational Linguistics (Volume 1: Long Papers)}, pages 944--964, St. Julian{'}s, Malta. Association for Computational Linguistics.

\bibitem[{Wu et~al.(2023)Wu, He, Liu, Sun, Liu, Han, and Tang}]{wu2023brief}
Tianyu Wu, Shizhu He, Jingping Liu, Siqi Sun, Kang Liu, Qing-Long Han, and Yang Tang. 2023.
\newblock A brief overview of chatgpt: The history, status quo and potential future development.
\newblock \emph{IEEE/CAA Journal of Automatica Sinica}, 10(5):1122--1136.

\bibitem[{Xiao and Wang(2021{\natexlab{a}})}]{xiao2021hallucination}
Yijun Xiao and William~Yang Wang. 2021{\natexlab{a}}.
\newblock On hallucination and predictive uncertainty in conditional language generation.
\newblock \emph{arXiv preprint arXiv:2103.15025}.

\bibitem[{Xiao and Wang(2021{\natexlab{b}})}]{xiao-wang-2021-hallucination}
Yijun Xiao and William~Yang Wang. 2021{\natexlab{b}}.
\newblock \href {https://doi.org/10.18653/v1/2021.eacl-main.236} {On hallucination and predictive uncertainty in conditional language generation}.
\newblock In \emph{Proceedings of the 16th Conference of the European Chapter of the Association for Computational Linguistics: Main Volume}, pages 2734--2744, Online. Association for Computational Linguistics.

\bibitem[{Xu et~al.(2023)Xu, Liu, Culhane, Pertseva, Wu, Semnani, and Lam}]{xu-etal-2023-fine}
Silei Xu, Shicheng Liu, Theo Culhane, Elizaveta Pertseva, Meng-Hsi Wu, Sina Semnani, and Monica Lam. 2023.
\newblock \href {https://doi.org/10.18653/v1/2023.emnlp-main.353} {Fine-tuned {LLM}s know more, hallucinate less with few-shot sequence-to-sequence semantic parsing over {W}ikidata}.
\newblock In \emph{Proceedings of the 2023 Conference on Empirical Methods in Natural Language Processing}, pages 5778--5791, Singapore. Association for Computational Linguistics.

\bibitem[{Yang et~al.(2023)Yang, Sun, and Wan}]{yang-etal-2023-new-benchmark}
Shiping Yang, Renliang Sun, and Xiaojun Wan. 2023.
\newblock \href {https://doi.org/10.18653/v1/2023.findings-emnlp.256} {A new benchmark and reverse validation method for passage-level hallucination detection}.
\newblock In \emph{Findings of the Association for Computational Linguistics: EMNLP 2023}, pages 3898--3908, Singapore. Association for Computational Linguistics.

\bibitem[{Yoon et~al.(2022)Yoon, Yoon, Yoon, Kim, and Yoo}]{yoon-etal-2022-information}
Sunjae Yoon, Eunseop Yoon, Hee~Suk Yoon, Junyeong Kim, and Chang Yoo. 2022.
\newblock \href {https://doi.org/10.18653/v1/2022.emnlp-main.280} {Information-theoretic text hallucination reduction for video-grounded dialogue}.
\newblock In \emph{Proceedings of the 2022 Conference on Empirical Methods in Natural Language Processing}, pages 4182--4193, Abu Dhabi, United Arab Emirates. Association for Computational Linguistics.

\bibitem[{Zhang et~al.(2023{\natexlab{a}})Zhang, Li, Das, Malin, and Kumar}]{zhang-etal-2023-sac3}
Jiaxin Zhang, Zhuohang Li, Kamalika Das, Bradley Malin, and Sricharan Kumar. 2023{\natexlab{a}}.
\newblock \href {https://doi.org/10.18653/v1/2023.findings-emnlp.1032} {{SAC}$^3$: Reliable hallucination detection in black-box language models via semantic-aware cross-check consistency}.
\newblock In \emph{Findings of the Association for Computational Linguistics: EMNLP 2023}, pages 15445--15458, Singapore. Association for Computational Linguistics.

\bibitem[{Zhang et~al.(2023{\natexlab{b}})Zhang, Qiu, Guo, Deng, Zhang, Zhang, Zhou, Wang, and Fu}]{zhang-etal-2023-enhancing-uncertainty}
Tianhang Zhang, Lin Qiu, Qipeng Guo, Cheng Deng, Yue Zhang, Zheng Zhang, Chenghu Zhou, Xinbing Wang, and Luoyi Fu. 2023{\natexlab{b}}.
\newblock \href {https://doi.org/10.18653/v1/2023.emnlp-main.58} {Enhancing uncertainty-based hallucination detection with stronger focus}.
\newblock In \emph{Proceedings of the 2023 Conference on Empirical Methods in Natural Language Processing}, pages 915--932, Singapore. Association for Computational Linguistics.

\bibitem[{Zhang et~al.(2023{\natexlab{c}})Zhang, Li, Cui, Cai, Liu, Fu, Huang, Zhao, Zhang, Chen et~al.}]{zhang2023siren}
Yue Zhang, Yafu Li, Leyang Cui, Deng Cai, Lemao Liu, Tingchen Fu, Xinting Huang, Enbo Zhao, Yu~Zhang, Yulong Chen, et~al. 2023{\natexlab{c}}.
\newblock Siren's song in the ai ocean: a survey on hallucination in large language models.
\newblock \emph{arXiv preprint arXiv:2309.01219}.

\bibitem[{Zhang et~al.(2019)Zhang, Han, Liu, Jiang, Sun, and Liu}]{zhang-etal-2019-ernie}
Zhengyan Zhang, Xu~Han, Zhiyuan Liu, Xin Jiang, Maosong Sun, and Qun Liu. 2019.
\newblock \href {https://doi.org/10.18653/v1/P19-1139} {{ERNIE}: Enhanced language representation with informative entities}.
\newblock In \emph{Proceedings of the 57th Annual Meeting of the Association for Computational Linguistics}, pages 1441--1451, Florence, Italy. Association for Computational Linguistics.

\bibitem[{Zhao et~al.(2023)Zhao, Nguyen, and Daum{\'e}~III}]{zhao-etal-2023-hallucination}
Lingjun Zhao, Khanh Nguyen, and Hal Daum{\'e}~III. 2023.
\newblock \href {https://doi.org/10.18653/v1/2023.findings-emnlp.266} {Hallucination detection for grounded instruction generation}.
\newblock In \emph{Findings of the Association for Computational Linguistics: EMNLP 2023}, pages 4044--4053, Singapore. Association for Computational Linguistics.

\bibitem[{Zhou et~al.(2020)Zhou, Neubig, Gu, Diab, Guzman, Zettlemoyer, and Ghazvininejad}]{zhou2020detecting}
Chunting Zhou, Graham Neubig, Jiatao Gu, Mona Diab, Paco Guzman, Luke Zettlemoyer, and Marjan Ghazvininejad. 2020.
\newblock Detecting hallucinated content in conditional neural sequence generation.
\newblock \emph{arXiv preprint arXiv:2011.02593}.

\end{thebibliography}
\bibliographystyle{acl_natbib}

\section{Appendix}

\subsection{Social Perspectives on Hallucinations} \label{appendix-social}
The exploration of hallucination in NLP is solely technocentric; however, its conceptual roots and applications are deeply intertwined with societal interpretations. To gain a better understanding of the term `hallucination,' it is important to consider its broader usage and implications beyond NLP. 
Hallucination has been studied across disciplines like psychology and neurology \cite{steele2017hallucination, legault2020encyclopedia}. Essentially, hallucinations involve \textit{``perceptions arising in the absence of any external reality – seeing or hearing things that are not there''} \cite{steele2017hallucination}. 
Although a version of this definition is commonly used in NLP, often with negative connotations, hallucinations have a wide scope, originating from fields such as neurology, and philosophy.

\textbf{Hallucination and Medicine:} 
Hallucination is believed to have neurological origins, often emerging from induced states such as drug usage, psychosis, sensory deprivation, or migraines \cite{legault2020encyclopedia}. These experiences can encompass various sensory modalities like auditory, visual, olfactory, tactile, gustatory, or somatic sensations \cite{boeving2020transpersonal}. 
Modern neurological research like \citet{legault2020encyclopedia} suggests that while hallucinations may not align with external reality, they are linked to brain regions responsible for processing perceptions from the external world.


\textbf{Hallucination and Creativity:}
Studies exploring hallucination in the context of creativity suggest that individuals with mild hallucinatory experiences may demonstrate enhanced generative creativity \cite{fink2014creativity,mason2021spontaneous}. Another prevalent notion is the use of hallucinations as a gateway to accessing intuition, creativity, and novel modes of thinking \cite{mason2021spontaneous}. However, there is a call for greater empirical rigor to establish robust connections between specific mental states leading to hallucinations and the creative thinking process \cite{fink2014creativity}. 

The analysis of differing perspectives on hallucination reveals its diverse interpretations, challenging prevalent assumptions within NLP. However, using the term 'hallucination' without its social context can foster misconceptions. Firstly, the `hallucinations' in AI systems result from discrepancies in input data and prompts rather than an absence of external senses. Secondly, this metaphor risks perpetuating stigma by linking negative AI phenomena with specific mental illness aspects \cite{pal-etal-2023-med}, potentially hindering destigmatization efforts in mental health domains \cite{maleki2024ai}. Lastly, given the widespread use of machine learning models, especially in medical fields \cite{ji2023towards}, a limited grasp of 'hallucination' context may lead to terminology misinterpretations.

\subsection{Hallucination in each NLP Task} \label{appendix-NLPtask}

\begin{table*}[]
\footnotesize
\centering
\begin{tabular}{|c|c|c|}
\hline
\textbf{Application} & \textbf{Definitions} & \textbf{Frequency}\\ \hline
Conversational AI & \begin{tabular}[c]{@{}c@{}}AI systems designed for natural language conversations, \\ understanding inputs, and generating appropriate responses\end{tabular} & 38\\ \hline
Abstractive Summarization &  \begin{tabular}[c]{@{}c@{}}Generating concise summaries by preserving main ideas\\ and context, often creating new sentence\end{tabular} & 16\\ \hline
Data2Text Generation & \begin{tabular}[c]{@{}c@{}}Automatically converting structured data into human-\\readable text, used in reporting and narratives\end{tabular} & 14\\ \hline
Machine Translation & \begin{tabular}[c]{@{}c@{}}Automatically translating text between languages using\\ computational methods like neural networks\end{tabular} & 12 \\ \hline
Image and Video Captioning & \begin{tabular}[c]{@{}c@{}}Generating descriptive captions for visual content, aiding\\ accessibility and understanding\end{tabular} & 8\\ \hline
Data Augmentation & \begin{tabular}[c]{@{}c@{}}Techniques to increase data diversity and quality, improving\\ model performance individual aspects of an entity\end{tabular} & 8\\ \hline
Miscellaneous & \multicolumn{1}{l|}{\begin{tabular}[c]{@{}c@{}}Encompasses additional non-accomodated tasks like natural\\ language inference and factuality detection\end{tabular}} & 7  \\ \hline
\end{tabular}
\caption{Frameworks of Sentiment and corresponding definitions in Sentiment Analysis} \label{table:application}
\end{table*}

We now analyze what aspects of the definitions of hallucination most commonly occur within each of our identified sub-fields of NLP\footnote{The breakdown of all the works associated with each of the subfields is in our \textit{Appendix}.} (Table \ref{table:application}).

\textbf{Conversational AI:} In this sub-field, hallucination encompasses fluency, non-factuality, and both intrinsic and extrinsic hallucinations. The definitions' facets highlight that dialogue systems must balance conversational fluency with factual consistency, aligning both with prior conversation and real-world truths.

\textbf{Abstractive Summarization:} Works in this sub-field mainly focuses on extrinsic and intrinsic hallucinations in defining it. Some definitions also mention the faithfulness of the generation. Despite the challenges of aligning with real-world facts and source consistency, prioritizing alignment and adherence to the original material has been shown to be essential in these works.

\textbf{Data2Text Generation:} Hallucinations are classified into extrinsic and intrinsic types, similar to abstractive summarization. Here, alignment with the underlying data is emphasized as the more critical factor when compared to the language used in generating the text.

\textbf{Machine Translation:}  Definitions of hallucination predominantly concentrate on extrinsic hallucination, with rare mentions of intrinsic hallucinations. This observation suggests a lesser concern for stylistic nuances in text generation within this field, with a greater emphasis on comprehending and conveying translated content accurately.

\textbf{Image and Video Captioning:} Models are expected to maintain consistency with the source while also incorporating real-world knowledge to address gaps and apply common sense. Consequently, the definition of hallucination in this context encompasses intrinsic, extrinsic, and non-factual elements, highlighting these requirements.

\textbf{Data Augmentation:}: Works from this domain often omit explicit definitions of hallucination, indicating a divergence in emphasis or a nascent exploration of this construct within this sub-field.

\textbf{Miscellaneous:} Encompassing tasks such as language inference and factuality detection, this category's definitions of hallucination encompass aspects like factuality, intrinsic and extrinsic hallucination, fidelity, and nonsensicality. It's evident that within these subfields, hallucination addresses both the stylistic aspects of model output and the fidelity and accuracy of generated content.

From the analysis of different subfields, it is evident that each perceives hallucination differently, emphasizing specific attributes such as factuality, fidelity, or linguistic styles like confidence, while potentially overlooking others. This diversity indicates that hallucination as a concept is still in its early stages in the field, with various frameworks emerging and a general lack of consensus regarding its definition and application. 
Furthermore, the lack of social aspects in hallucination discussions in these subfields contrasts with the broader understanding and research in fields like healthcare.

\subsection{Supplementary Survey Analysis} \label{appendix-survey}

\subsubsection{Weaknesses of LLM}
Before delving into inquiries about hallucinations in LLMs, it is crucial to gain insights into the perceived weaknesses of these models from the participants' perspective, as well as understand how frequently they utilize these models in their work.  

The survey results indicate that a significant portion of researchers heavily utilize LLMs in their daily life. Specifically, 67.28\% of respondents reported using these models atleast once a day, while 20.37\% mentioned using them all the time, highlighting the ubiquity of these models. 

Upon analyzing the themes derived from participants' responses on the \textit{weaknesses of generative AI tool}s, it was observed that a substantial majority (55\%) of researchers perceive the main weaknesses to be the generation of misinformation and hallucinations, despite both phenomena being essentially similar in nature. For instance,


\textit{\say{I have been exploring these models to see what they get right and wrong. They get a lot of things wrong -- what some people call ``hallucinations''.}}---Emeritus Professor, NLP

Some of the other important weaknesses mentioned by the respondents are: biases, not following the prompts correctly, complex language, and not having a long memory. For example,

\textit{\say{They produce a lot of inaccurate replies with great confidence. These models also tend to be very biased toward many socio-demographic groups.}}---Graduate student, GenerativeAI

\textit{\say{It is hard to distinguish whether the information provided by them is accurate or not. Sometimes, the models generate text with reasoning making it sound convincing enough to be true - but ends up being incorrect ultimately.}}---Industry, GenerativeAI

The responses highlight a critical concern within the research community regarding the reliability and accuracy of outputs generated by LLMs, with potential implications for various applications and domains, providing us with a strong motivation behind this study. 

The widespread use of LLMs, particularly prominent models such as GPT 3, 3.5, and 4, highlights their importance and impact on research and industry practices. However, it's noteworthy that respondents also mentioned other LLM models that they use or are familiar with. These include Mistral, BERT, LLaMA2, Midjourney, ClaudeAI, Gemini, Vicuna, t5, Falcon, PaLM, Imagen, Dolly, Perplexity, among others.

\subsubsection{Social Ramifications of Hallucination}
Participants were prompted to explain the effects of hallucination on their work/daily life. 
The resulting themes, from our qualitative analysis of their inputs, are outlined below:

\textbf{Not Good for Education:}
Respondents raised concerns about the extensive use of these models by students for homework, indicating potential negative impacts on their performance and learning abilities. The respondents believe that such reliance on these models can lead to a degradation in students' learning. Additionally, respondents express skepticism about the suitability of these models for checking homework assignments.

\textit{\say{I don't actually use AI for my work; I just want to be aware of what it can do because my students are probably using it for their homework.  It could have an impact on students' mastery of the material.}}---Associate Prof, Biotechnology

\textbf{Not Good for Scholarly Work:}
Several respondents noted that these models are not effective for scholarly purposes, citing instances where the models generated information that was not present in the original paper. They express concerns that if researchers rely on these models for tasks like literature summarizing, it could lead to a deterioration in scholarly processes. For example:

\textit{\say{They tend to generate a lot of misinformed facts about certain groups or cultures that I have seen happen often. They also generate 'facts' from scholarly works where the papers would not have mentioned the same.}}---Graduate student, NLP


\textbf{Struggle with Code Generation:}
The models were deemed inefficient for code generation by multiple respondents, often producing code that lacks utility due to hallucinations. Respondents highlighted mismatches between the generated code and its intended purpose, emphasizing the need for thorough review before utilization. Various concerns were raised, including the loss of context during prolonged interactions, inaccuracies in complex coding tasks leading to erroneous outputs, fabrication of functions or attributes, inaccuracies in both code and associated theoretical concepts, necessitating extensive debugging and corrections, and a tendency to cycle back to previously incorrect suggestions despite error notifications. 

\textit{\say{I was asking an AI to generate me a piece of code. It ended up picking some code from one website and some from another and combining it. However those two websites (they were linked by chatgpt) we're using different versions of the library so the resulting code couldn't be executed.}}---Industry, Network and Security

\textbf{Increase in Time for Task:}
A common sentiment among respondents is that these models frequently produce errors or false information, resulting in potential time wastage. While they acknowledge occasional helpfulness, there's a consensus that reliance on these models can often lead to unfavorable outcomes, particularly when verifying outputs. This dependency on verification contributes to increased task duration, adding extra work and time toward the project's conclusion, as noted by several respondents.

\textit{\say{I use GPT API to conduct analysis for some of my work and accuracy and consistency would be good in my context, and I have to find ways to finetune it before I can trust the results of the analysis, which added more work on my end.}}---Graduate Student, HCI

\textbf{Misleading and Distrust:}
Generating incorrect outputs with confidence can lead to the dissemination of non-existent knowledge, such as misleading information in the literature that may confuse individuals with incorrect concepts. Most of our respondents mentioned this concern. Moreover, it poses challenges in differentiating between accurate AI responses and hallucinations, particularly for users lacking expertise in the relevant subject matter.

\textit{\say{It leads to problems if even I do not have any idea about the work. It is hard to differentiate if it is a genuine output or hallucination.}}---Graduate Student, Data Science

\begin{figure}[h]
  \centering
  \includegraphics[scale = 0.3]{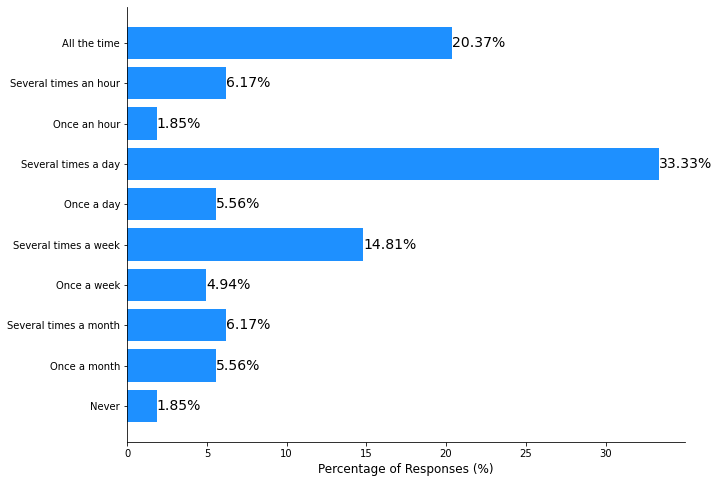}
  \caption{Frequency of Text Generation Model Usage}
  \label{fig:Use}
\end{figure}

\subsubsection{An External Viewpoint} 

Additionally, our survey of 51 researchers who do not specialize in AI revealed that all except 3 have used text-generation models like various versions of ChatGPT. Despite their fields not being directly related to AI, a significant number integrate these tools into their workflow, with 19.6\% using them multiple times daily and 11.76\% using them several times per hour. Their extensive usage has allowed them to identify several limitations in the models; they are: \textit{mathematical inaccuracy, outdated information, misinformation, poor performance with complex tasks and creative thinking, lack of specificity in-depth, overconfidence, lack of transparency, bias,} and \textit{irrelevant responses.}

Based on the definitions provided, it is observed that there is a lack of clarity among the respondents regarding what constitutes a `hallucination' in generative AI models, with perspectives varying widely. Thematic analysis of their responses indicates that the predominant view associates 'hallucination' with \textit{the generation of nonfactual content and misinformation by AI systems}. That means these models are generating facts that are not real and misleading. The remaining themes are f\textit{actually incorrect, biased outputs, incompleteness, misinformation with confidence}, and \textit{nonsensical but good-looking texts}.

The results demonstrate the unclear comprehension and significance attributed to hallucination in LMs beyond the field of NLP and AI. There is a pressing need to enhance public understanding of the concept of hallucination, emphasizing its meaning and strategies for mitigation. Given the increasing prominence of LMs as sociotechnical systems \cite{narayanan2023towards}, it is crucial to grasp their social interactions and potential societal ramifications.

\subsubsection{Additional Impacts and Concerns} 
We analyzed perceptions when participants were asked about any additional concerns during the survey. Participants emphasized the necessity for greater control and more nuanced mechanisms to address and manage AI hallucinations effectively. Presently, the detection and rectification of hallucinations rely heavily on meticulous human review, highlighting the need for tools designed specifically to identify and mitigate such occurrences. The presence of hallucinations can significantly impact the credibility and acceptance of generative models among the general public. These issues arise due to the inherent limitations of generative algorithms and the absence of access to real-time external knowledge.

Transparency regarding the limitations of generative AI is deemed essential through our findings, and user education is seen as a key factor in mitigating risks associated with the unchecked use of AI-generated content, as the responsibility for identifying hallucinations often falls on the user. While inaccuracies in non-critical applications, like movie suggestions, may be tolerable, according to our survey, they are deemed crucially problematic in contexts such as business decision-making, law, or health \cite{Dahl2024legal}.

\subsection{Survey Questions} \label{appendix_questions}
In this section, we provide the content and the questions that were presented in the survey:

\textbf{Survey Title:} Insights of Usage and Issues with Text Generative Models and Tools

\begin{enumerate}
  \item How did you receive the survey? \textit{(Social Media Posts, Direct email, Direct messages, Others)}
  \item What is your current country of residence? \textit{(Open-ended)}
  \item What sector do you associate with? \textit{(Academia, Industry, Others)}
  \item What is your field of expertise? \textit{(Open-ended)}
  \item Does your research work directly involve studying or developing Artificial Intelligence (AI)? \textit{(Yes, No)}
  \item How often do you use Text generation models (like ChatGPT/Gemini)? \textit{(All the time, Several times an hour, once an hour, several times a day, Once a day, Several times a week, once a week, Several times a month, Once a month, Never)}
  \item Which text generation models have you used, if any? \textit{(Open-ended)}
  \item What weaknesses do you perceive in the models that you have used(if any)? \textit{(Open-ended)}
  \item Are you familiar with the concept of 'hallucinations' in AI-generated text? \textit{(Extremely familiar, Very familiar, Moderately familiar, Slightly familiar, Not at all familiar)}
  \item What, according to you, is 'hallucination' in generative AI models?\textit{(Open-ended)}
  \item Do you consider 'hallucinations' to be a weakness when using these models? \textit{(Yes, No)}
  \item How frequently do you encounter that text generation models produce 'hallucinated' content that is factually incorrect or unrelated to the input? \textit{(Very frequently, frequently, Occasionally, rarely never)}
  \item If you have an alternate term in mind to describe the phenomenon instead of 'hallucination' (e.g., fabrications, confabulations, etc.), kindly specify it along with an explanation(Mention NA if none). \textit{(Open-ended)}
    \item Can you provide an example where a hallucination in text generation had or can have an impact on your work (Mention NA if None)? \textit{(Open-ended)}
    \item Do you have any additional comments or insights regarding the hallucination? (if any) \textit{(Open-ended)}
\end{enumerate}

\subsection{Works and Application}
We illustrate the examples and categories of works that were looked into for understanding the various applications of hallucinations. We categorize the research on hallucinations into 7 major categories. The definitions and categories of all the applications are mentioned in Table \ref{table:application}.

\textbf{Abstractive Summarization:} \citet{zhang-etal-2019-ernie, son-etal-2022-harim, maynez-etal-2020-faithfulness, choubey-etal-2023-cape, cao2021hallucinated, marfurt-henderson-2022-unsupervised, akani2023reducing, van-der-poel-etal-2022-mutual, chen2023fidelity, dong-etal-2022-faithful, shen-etal-2023-mitigating, nan-etal-2021-entity, chen-etal-2021-improving, ladhak2023pre, nan-etal-2021-entity, flores-cohan-2024-benefits}

\textbf{Conversational AI:} \citet{liu-etal-2022-token, zhou2020detecting, ji-etal-2023-rho, zhang-etal-2023-enhancing-uncertainty, yang-etal-2023-new-benchmark, das-etal-2022-diving, bouyamourn-2023-llms, sun-etal-2023-towards, sadat-etal-2023-delucionqa, slobodkin2023curious, ramakrishna2023invite, xiao-wang-2021-hallucination, shuster-etal-2021-retrieval-augmentation, nie-etal-2019-simple, longpre-etal-2021-entity, dziri2022faithdial, maheshwari-etal-2023-open, ladhak2022contrastive, xu-etal-2023-fine, chen2023exploring, goldberg2022findings, sundar-heck-2023-ctbls, roller-etal-2021-recipes, mielke2022reducing, roller-etal-2021-recipes, massarelli-etal-2020-decoding, weller-etal-2024-according, smith-etal-2022-human}

\textbf{Data Augmentation:} \citet{jian-etal-2022-embedding, ji-etal-2023-rho, friedl-etal-2021-uncertainty, samir-silfverberg-2022-one, anastasopoulos2019pushing, narayanan2023unmasking} 

\textbf{Image and Video Captioning:} \citet{xiao-wang-2021-hallucination, dai-etal-2023-plausible, rohrbach-etal-2018-object, li-etal-2023-evaluating, testoni-bernardi-2021-ive, son-etal-2022-harim, dai-etal-2023-plausible, li-etal-2023-evaluating, liu-wan-2023-models}

\textbf{Machine Translation:} \citet{wang-sennrich-2020-exposure, raunak-etal-2021-curious, dale2022detecting, guerreiro-etal-2023-optimal, xu-etal-2023-fine, pfeiffer-etal-2023-mmt5, guerreiro-etal-2023-looking, dale-etal-2023-halomi, irvine2014hallucinating, ferrando-etal-2022-towards, vu-etal-2022-overcoming, muller-etal-2020-domain, waldendorf-etal-2024-contrastive}

\textbf{Data2Text Generation:} \citet{gonzalez-corbelle-etal-2022-dealing, shi-etal-2023-hallucination, yoon-etal-2022-information, filippova-2020-controlled, kothyari-etal-2023-crush4sql, lango-dusek-2023-critic, cirik2022holm, fei-etal-2023-scene, obaid-ul-islam-etal-2023-tackling, qiu-etal-2023-detecting, testoni-bernardi-2021-ive, gonzalez2022dealing, islam2023tackling, polat2023improving}

\textbf{Miscellaneous:} \citet{manakul-etal-2023-selfcheckgpt, ji-etal-2023-towards, maharaj-etal-2023-eyes, mckenna-etal-2023-sources, pal-etal-2023-med, zhao-etal-2023-hallucination, berbatova-salambashev-2023-evaluating, wu-etal-2024-lamini}

\end{document}